\documentclass[10pt,journal,compsoc]{IEEEtran}
\ifCLASSOPTIONcompsoc
  \usepackage[nocompress]{cite}
\else
  \usepackage{cite}
\fi
\ifCLASSINFOpdf
   \usepackage[pdftex]{graphicx}
\else
  \usepackage[dvips]{graphicx}
\fi
\usepackage{amsmath}
\usepackage{amsfonts}
\usepackage{algorithm}
\usepackage{algorithmic}

\usepackage{array}

\ifCLASSOPTIONcompsoc
  \usepackage[caption=false,font=footnotesize,labelfont=sf,textfont=sf]{subfig}
\else
  \usepackage[caption=false,font=footnotesize]{subfig}
\fi
\begin{document}
%
\title{Fast and Globally Optimal Rigid Registration of 3D Point Sets by Transformation Decomposition}
%
%
%
%

\author{Xuechen~Li, Yinlong~Liu,
        Yiru~Wang, Chen~Wang, Manning~Wang,
        and~Zhijian~Song
\IEEEcompsocitemizethanks{\IEEEcompsocthanksitem Xuechen~Li, Yiru~Wang, Chen~Wang, Manning~Wang, and~Zhijian~Song are with Digital Medical Research Center, School of Basic Medical Sciences, Fudan University, Shanghai, 200032, China.\protect\\
E-mail: $\{$lixuechen, yiruwang18, wangchen17, mnwang, zjsong$\}$@fudan.edu.cn
}
\IEEEcompsocitemizethanks{\IEEEcompsocthanksitem Yinlong~Liu is with Department of Informatics, Technical University of Munich, Munich, 385748, Germany.\protect\\
E-mail: yinlong.liu@tum.de
}
\thanks{(Corresponding authors: Manning Wang and Zhijian Song.)}}

\IEEEtitleabstractindextext{%
\begin{abstract}
The rigid registration of two 3D point sets is a fundamental problem in computer vision. The current trend is to solve this problem globally using the BnB optimization framework. However, the existing global methods are slow for two main reasons: the computational complexity of BnB is exponential to the problem dimensionality (which is six for 3D rigid registration), and the bound evaluation used in BnB is inefficient. In this paper, we propose two techniques to address these problems. First, we introduce the idea of translation invariant vectors, which allows us to decompose the search of a 6D rigid transformation into a search of 3D rotation followed by a search of 3D translation, each of which is solved by a separate BnB algorithm. This transformation decomposition reduces the problem dimensionality of BnB algorithms and substantially improves its efficiency. Then, we propose a new data structure, named 3D Integral Volume, to accelerate the bound evaluation in both BnB algorithms. By combining these two techniques, we implement an efficient algorithm for rigid registration of 3D point sets. Extensive experiments on both synthetic and real data show that the proposed algorithm is three orders of magnitude faster than the existing state-of-the-art global methods.
\end{abstract}

}

\maketitle

\IEEEdisplaynontitleabstractindextext

%
\IEEEpeerreviewmaketitle

\IEEEraisesectionheading{\section{Introduction}\label{sec:introduction}}

%
%
%
%
\IEEEPARstart{T}{he} rigid registration of two 3D point sets is performed to find a spatial transformation in \emph{SE}(3) to best align the two point sets. Rigid registration is an essential component of computer vision tasks such as object detection and recognition [1], [2], robot navigation [3], [4], [5] and model construction from multiple partial scans [6], [7]. When at least four pairs of corresponding points are known, closed-form solutions exist for calculating the 6-DOF rigid transformation, which includes a translation in  $\mathbb{R}^3$ and a rotation in \emph{SO}(3) [8]. If the putative correspondences have outliers, the problem can still be approached in a RANSAC framework [9], [10], [11], [12]. However, the problem becomes more difficult when the point correspondences are not known a priori [13], which is usually known as the simultaneous pose and correspondence problem (SPC problem) [14].

The SPC problem was first solved locally, and the Iterative Closest Point (ICP) [15] method proposed by Besl and McKay in 1992 can be regarded as a milestone in this area. ICP starts with an initial guess of the transformation between the two point sets and then iterates between finding the correspondence under the current transformation and updating the transformation with the newly found correspondence. ICP is widely used because it is rather straightforward and easy to implement in practice; however, its biggest problem is that it does not guarantee finding the globally optimal transformation. In fact, ICP converges within a very small basin in the parameter space, and it easily becomes trapped in local minima. Therefore, the results of ICP are very sensitive to the initialization, especially when high levels of noise and large proportions of outliers exist. To enlarge the convergence basin and alleviate local minima problems, many variants of ICP have been proposed, including [16], [17], [18], [19], [20], [21], [22], [23], [24]. These variants resemble classical ICP in their alignment criterion, which is based on the distance between corresponding points, and their alternating optimization strategy between the transformation and the correspondence. Thus, their improvements regarding the problem of local convergence are limited. Another line of research models point sets as probability distributions and transforms point set registration into a problem of aligning two probability distributions; these models include KC [25], GMM [26], [27], CPD [28],and [29], [30], [31], and so on. The best alignment is achieved by minimizing the statistical difference between the two probability distributions. The objective function under these models can be made smoother and involve fewer local minima by choosing proper kernels when constructing the probability distributions from the original point sets to enlarge the convergence basin. However, these models are still unable to guarantee global optimality; consequently, they still require a good initialization to avoid converging to a totally wrong transformation.

Recently, there has been a surge of studies on global point set registration methods that theoretically guarantee finding the global optimal transformation regardless of the initialization. Breuel’s [32] is one of the earliest works to achieve global optimality in point set registration, and was followed by [14], [33], [34], [35], [36], [37], [38] and [39]. All these global methods utilize the Branch-and-Bound (BnB) optimization framework to search for the global solution. Although BnB guarantees global optimality, the existing methods exhibit excessive runtimes, which greatly compromises their practicability. The low time efficiency of these methods occurs mainly for the following two reasons.

First, the time complexity of BnB methods is exponential in the dimensionality of the parameter space, and the 6-DOF parameter space of \emph{SE}(3) in 3D rigid point set registration is apparently too high for BnB. Three schemes exist to search the 6-DOF parameter space. The first approach is to search the 6D space directly, which is extremely time-consuming. GOGMA [34] explored this scheme, which required the assistance of GPUs to complete registration in a reasonable amount of time. The second scheme employs a structure called nested BnB, in which an outer BnB and an inner BnB are used to search the translation and the rotation, respectively. This approach was proposed in Go-ICP and was also adopted by [39], [40]. However, the improvement was limited: both algorithms still required thousands of seconds to register even moderately sized point sets. The third approach is to decompose the 6D parameter space of rigid transformation into two separate lower-dimensional spaces, 3D rotation and 3D translation (e.g., [41], [36], [42]) and then search for the rotation and translation separately in these two 3D spaces. This approach is promising, but [36] requires a GPU to achieve decomposition and [42] does not guarantee the global optimality of the rotation.

Second, the low efficiency of the bounds used in current BnB-based global registration methods is another bottleneck of the overall algorithm. Bound efficiency has two aspects, namely, its tightness, which determines how many iterations the algorithm will need to converge, and the time needed to evaluate the bound in each branch, which determines the calculation time of each iteration. Almost every new global method using BnB optimization derives a new bound, and much focus has been placed on making these bounds tighter. A series of works [33], [34], [38], [39], [43], for example, attempted to derive bounds that were increasingly geometrically tight. However, the efficiency with which the bound can be evaluated also plays a considerable role in the overall runtime of a BnB algorithm. Because the evaluation of bounds usually involves calculations between each pair of points in the two sets, whose computational complexity is usually $O(N^2)$, where $N$ is the number of points in each set. Go-ICP [33] utilizes distance transform (DT) to reduce the computational complexity of bound evaluation from $O(N^2)$ to $O(N)$, which greatly improves its speed but loses the theoretically guaranteed global optimality.

In this paper, we propose novel approaches to address the above two problems and achieve a very efficient globally optimal method for 3D rigid point set registration three orders of magnitude faster than existing global methods and approximately 10 times faster than Go-ICP [33] which achieves its speed improvement through the distance transform. On one hand, we propose the idea of translation invariant vectors (TIVs), allowing us to first search the 3D rotation between the two point sets to be registered by matching these TIVs and then searching for the 3D translation between the two point sets given the known rotation between them. A toy 2D example to illustrate the construction and alignment procedures of TIVs is presented in Fig. 1. Therefore, the BnB search in 6D rigid transformation space is decomposed into a BnB search in the 3D rotation space followed by a BnB search in the 3D translation space. On the other hand, we propose using the $L_\infty$-based consensus set as the objective function for point set registration and propose a new data structure, named 3D Integral Volume (3DIV), to precisely and efficiently evaluate the bounds during the rotation and the translation searches.

\begin{figure}[tbp]
\centering
\includegraphics[width=3.5in]{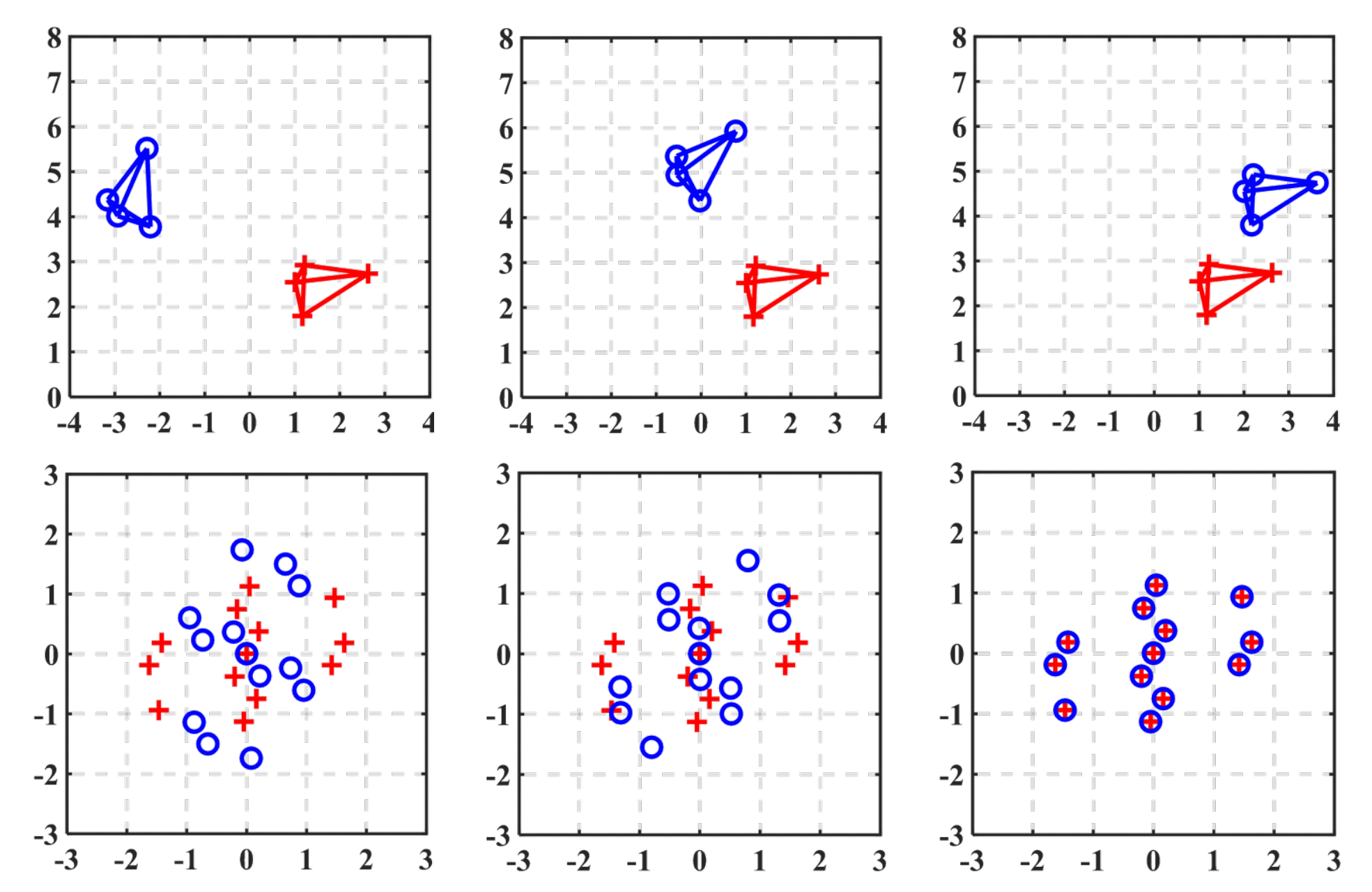}
\caption{2D illustration of constructing and aligning the TIVs, which are essentially the vectors connecting point pairs. In the first row, the three images from left to right, are the fixed point set (red) and the moving point set (blue) with relative rotations of 60°, 30° and 0, respectively. The solid lines denote the TIVs. In the second row, the three images from left to right, are the TIVs shown as data points, corresponding to the relative rotations in the first row. The blue circles denote the TIVs constructed from the moving point set and the red crosses denote the TIVs constructed from the fixed point set. The last column shows that the two sets of TIVs are aligned when there is no relative rotation between the two original point sets.}
\label{fig_sim}
\end{figure}

\section{Related Work}
In this paper, we focus on globally optimal registration of 3D point sets. Thus, in this section, we mainly review works that are highly relevant to global methods. For details about local methods, such as ICP, GMMReg, CPD and others, readers are referred to recent review papers [44], [45]. In addition, pure rotation search methods are briefly discussed because we propose a new pure 3D rotation search method that is embedded in our framework for rigid registration of 3D point sets.
\subsection{Global Point Set Registration Methods}
The need to obviate the local minima problem essentially prompted research on global registration methods. As one of the earliest global approaches, Breuel’s work [32] performed well for 2D registration, but its extension to 3D was nontrivial. Subsequently, many researchers explored the same idea of using BnB optimization framework to achieve global optimization. Li and Hartley [14] solved the 3D SPC problem by integrating the BnB framework with Lipschitz optimization, which provides a general way of deriving bounds; however, the bounds are fairly loose. In addition, this method assumed that the two point sets had the same size; consequently, it cannot deal with outliers or partially overlapping point sets, which are very common in practice. Olsson et al. [37] introduced a convex relaxation approach to solving the problem globally, but the relaxation requires some known point-to-plane, point-to-line or point-to-point correspondences, which are not available in many applications.

A series of recent studies on global point set registration explore the geometric bound of a point under rotation, which is based on the work of [13]. Go-ICP proposed by Yang et al. [33] was the first globally optimal algorithm that could perform rigid registration of two 3D point sets. In Go-ICP, the geometric bound is expressed as an uncertainty ball. Go-ICP [33] minimized the same   distance-based objective function as the original ICP algorithm and derived the lower bound of the minimum using the geometric bound. [33] designed a nested BnB pattern comprising an outer and an inner BnB. The outer BnB searches the rotation space of \emph{SO}(3), and the inner BnB searches for the corresponding optimal translations. Campbell and Petersson [34] registered two point sets by aligning the Gaussian mixture models (GMM) established from the original point sets. They derived a tighter geometric bound, which is a cap on the uncertainty ball used in [33], to establish the bound for their probability distribution-based objective function. The two GMMs with 50 components were generated by support-vector-parameterized Gaussian mixtures (SVGMs) [46], which are also time-consuming processes. In addition, GOGMA searches the 6D space of rigid transformation; therefore, it utilizes a GPU to accelerate the algorithm. Bustos et al. [39] utilized the same geometric bound as [34] but established an objective function based on a consensus set. The evaluation of the upper bound of the maximum consensus set is accelerated by stereographically projecting the cap-shaped geometric bound in the 3D space into a circle on a 2D plane. Essentially, Go-ICP, GOGMA and Glob-GM-ML [39] are all based on the geometric bound of a point under rotation developed in [13], while extending the bound to translation is trivial. We also use the geometric bound proposed in [13] to derive our bound in this paper, but we do not derive a tighter bound. Instead, we derive a slightly looser bound in the rotation search, which the proposed 3DIV data structure can evaluate much faster.

In addition to the geometric bounds, some works have explored other ways to develop BnB-based global methods and the required bounds. Lian and Zhang proposed Asymmetric Point Matching (APM) [35] to solve the point matching problem, which can be regarded as an equivalent form of point set registration. They reduce the original point matching problem to a concave quadratic assignment problem and propose an efficient lower bound to solve it with the BnB framework. Unlike the rigid registration methods, APM calculates the 12 DOF of 3D affine transformations. However, it assumes that each point in one set must have a counterpart in the other, which is rarely realistic in practice. Straub et al. [36] proposed a two-stage BnB algorithm using a data structure based on surface normals. As mentioned in Section 1, [36] adopted the scheme of decomposing the search of a 6D rigid transformation into separate searches for a 3D rotation and a 3D translation. This scheme is more efficient than the directly searching the 6D rigid transformation from the perspective of BnB optimization. However, the process of constructing the surface-normal-based data structure is time-consuming; therefore, it was implemented using GPUs. Our method resembles [36] in that it applies transformation decomposition, but we propose a more efficient method of transformation decomposition and also propose a new data structure, 3DIV, to accelerate the rotation and translation search.
\subsection{Rotation Search}
Although this paper does not focus on the pure rotation search problem, as a part of our decomposition-based registration framework, we also propose a 3D rotation search method based on consensus maximization and a novel geometric bound. Similarly, some other rigid registration methods provided a separable rotation kernel such as in [38], [39]. At the same time, some works have studied the pure rotation search problem [47], [48], [49]. All these rotation search methods [38], [39], [47], [48], [49] can be categorized into two groups: one group [47], [48], [49] exploits image-based feature matches, while the other [38], [39] operates on raw 3D points. In this context, our method belongs to the latter group. Matchlist was added to [38] in an extended version [39] to accelerate bound evaluation. To the best of our knowledge, MCIRC-ML in [39] achieves the best performance in terms of pure rotation search. In the experimental section, we show that our rotation search method is faster in our decomposition framework.
\subsection*{Contributions}
The main contributions of this paper can be summarized as follows:

1. We propose a new transformation decomposition framework for rigid registration of 3D point sets by introducing TIVs. We construct two sets of TIVs from the two original point sets. These two TIV sets have the following key characteristics: first, they can be aligned through pure rotation, and their relative rotation is the same as the relative rotation between the two original point sets; second, the TIVs are invariant to translations of the original point sets. In this way, we decompose the search of a 6D rigid transformation into a 3D rotation search followed by a 3D translation search when the rotation is known. This problem dimensionality reduction helps speed up the solution of this problem with the BnB optimization framework, whose computational complexity is exponential to the problem dimensionality.

2. We introduce a new objective function for the point set registration problem, which is based on the consensus set measured by the   distance, and we derive a new geometric bound suitable for evaluating the bound for this objective function.

3. We propose a new, faster 3DIV structure to accelerate the evaluation of bounds. The time cost of 3DIV construction is extremely small.

4. We implement a fast global optimal algorithm for rigid registration of two 3D point sets. Extensive experiments on both synthetic and real data show that our algorithm is three orders of magnitude faster than the state-of-the-art global methods and approximately 10 times faster than Go-ICP accelerated by DT.

The rest of this paper is organized as follows: we introduce the idea of using TIVs and the decomposition of the 6-DOF rigid transformation in Section 3. After decomposition, the efficient rotation and translation search methods are presented in Section 4 and Section 5, respectively. In Section 4, we also introduce our $L_\infty$ distance-based objective function and the faster 3DIV structure. In Section 6, we evaluate the proposed algorithm for rigid registration of 3D point sets and compare it with state-of-the-art global methods. Section 7 concludes the paper.

\section{Decomposition of Rigid Transformation}
In this section, we first give the formulation of the 3D rigid point set registration problem and then describe how to decompose the 6-DOF problem into two 3-DOF sub-problems using the proposed translation invariant vectors (TIVs).

\subsection{ Problem Formulation of 3D Rigid Point Set Registration}
Given two 3D point sets, $\mathcal{M}=\{\textbf{m}_{i}\}{^M_{i=1}}$ and $\mathcal{S}=\{\textbf{s}_{j}\}{^N_{j=1}}$, where $\textbf{m}_{i}\in{\mathbb{R}^3}$ and $\textbf{s}_j\in{\mathbb{R}^3}$, the objective of rigid registration of the two point sets is to find a transformation $T_{opt}\in{SE(3)}$ such that
\begin{equation}
T_{opt}=\mathop{\arg\max}_{T\in{SE(3)}}Q(T(\mathcal{M}),\mathcal{S}).
\end{equation}

Here $\mathcal{M}$ is usually called the model set or moving set, and $\mathcal{S}$ is usually called the scene set or fixed set. The rigid transformation $T$ is composed of a rotation and a translation; therefore, for each point $\textbf{m}_i$ in $\mathcal{M}$, its coordinate after being transformed by $T$ is $\textbf{m}{'_i}=T(\textbf{m}_i)=\textbf{Rm}_i+\textbf{t}$, where $\textbf{R}$ is a 3D rotation matrix and $\textbf{t}$ is a 3D translation vector. The function $Q$ measures the alignment between $T(\mathcal{M})$ and $\mathcal{S}$. Different algorithms may define different functions to measure the alignment; two popular strategies are minimizing the sum of the $L_2$ distance between the corresponding points [15], [24] and maximizing the number of consensus points [32], [38], [39], in which a pair of points is considered a consensus when their $L_2$ distance is below a threshold. The advantage of the consensus strategy is that it is inherently robust to outliers. In this paper, we adopt the consensus set maximization strategy and use the $L_\infty$ distance to define the consensus. The objective function used in the rotation search and translation search in this paper and their fast global maximization algorithm are discussed in detail in Sections 4 and 5; in the following subsection, we first introduce the idea of how to decompose the search problem in the 6D space of rigid transformation into the problem of searching two 3D spaces (i.e., the 3D rotation space and the 3D translation space). This decomposition plays a central role in the efficiency of the proposed 6-DOF rigid point set registration algorithm.

\subsection{Transformation Decomposition by Translation Invariant Vectors}
As defined in Section 3.1, the rigid transformation $T$ has 6 DOF, and direct optimization over $T$ involves finding a 6D transformation parameter, which is slow using BnB-based global optimization methods. The basic idea of rigid transformation decomposition is straightforward. Instead of optimizing over the rigid transformation $T$ directly, we construct a set of features that are invariant to translation so that we can first optimize over 3D rotation to align these features. At the same time, the optimal rotation to align these features is also the optimal relative rotation between the two original point sets. After finding the relative rotation, another optimization over 3D translation can be conducted to recover the optimal translation. Thus, our method decomposes the 6-DOF optimization problem into two 3-DOF sub-problems. Here, we describe our decomposing strategy in detail.

Let $T^*$ be the optimal transformation from $\mathcal{M}$ to $\mathcal{S}$. Then, for a point $\textbf{m}_i$ in $\mathcal{M}$, if a corresponding point $\textbf{s}_j$ exists in $\mathcal{S}$, we have
\begin{equation}
\textbf{s}_j=\textbf{m}{'_i}=T^*(\textbf{m}_i)=\textbf{R}^*\textbf{m}_i+\textbf{t}^*,
\end{equation}
where $\textbf{R}^*$ and $\textbf{t}^*$ denote the optimal rotation and the optimal translation, respectively. Given any two points $\textbf{m}_{i1}$ and $\textbf{m}_{i2}$  in $\mathcal{M}$ and their corresponding points $\textbf{s}_{j1}$ and $\textbf{s}_{j2}$ in $\mathcal{S}$, we have $\textbf{s}_{j1}=\textbf{R}^*\textbf{m}_{i1}+\textbf{t}^*$ and $\textbf{s}_{j2}=\textbf{R}^*\textbf{m}_{i2}+\textbf{t}^*$. Let $\textbf{m}_{i1-i2}$ be the vector from $\textbf{m}_{i1}$ to $\textbf{m}_{i2}$ and $\textbf{s}_{j1-j2}$ be the vector from $\textbf{s}_{j1}$ to $\textbf{s}_{j2}$. Then, we have
\begin{equation}
\begin{split}
\textbf{s}_{j1-j2}=\textbf{s}_{j1}-\textbf{s}_{j2}&=(\textbf{R}^*\textbf{m}_{i1}+\textbf{t}^*)-(\textbf{R}^*\textbf{m}_{i2}+\textbf{t}^*)\\
&=\textbf{R}^*(\textbf{m}_{i1}-\textbf{m}_{i2})=\textbf{R}^*\textbf{m}_{i1-i2}.
\end{split}
\end{equation}

From (3), we can see that the feature vectors $\textbf{m}_{i1-i2}$ and $\textbf{s}_{j1-j2}$  can be aligned by the optimal rotation $\textbf{R}^*$ used to align the original point sets. In addition, both $\textbf{m}_{i1-i2}$ and $\textbf{s}_{j1-j2}$ are invariant to translations of the point sets. Therefore, by constructing these feature vectors, we can first find the optimal relative rotation $\textbf{R}^*$ by aligning these vectors without considering the relative translation between the two original point sets. The feature vectors (e.g., $\textbf{m}_{i1-i2}$ and $\textbf{s}_{j1-j2}$) are called translation invariant vectors (TIVs). After the optimal rotation  $\textbf{R}^*$  is found, it can be applied to the data points. Then, the rotationally aligned data points can be used to search for the optimal translation  $\textbf{t}^*$. Please note that a TIV is stored in the same way as a 3D point, so a TIV can be interchangeably considered as a 3D point or a 3D vector. Therefore, performing a rotation search on TIVs is essentially the same as performing a rotation search on 3D points. We outline our transformation decomposition-based 3D rigid registration framework in Fig. 2.

\begin{figure*}[!t]
\centering
\includegraphics[width=7in]{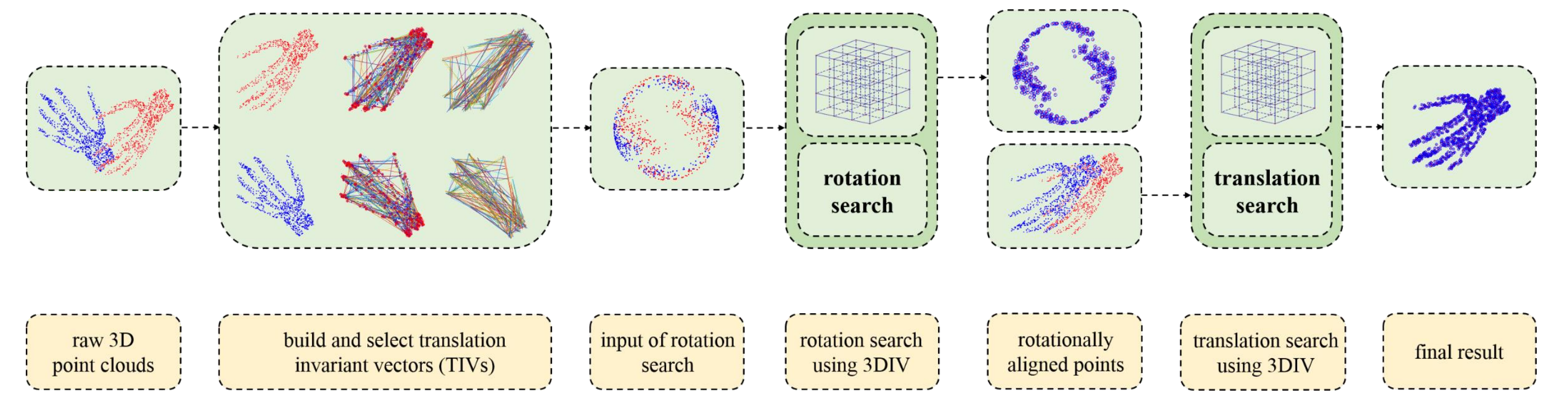}
\caption{Brief outline of our transformation decomposition-based 3D rigid point set registration method. From each of the two raw 3D point sets, a set of translation invariant vectors (TIVs) are constructed and a subset of the TIVs is selected to form the input of the rotation search. A global optimal rotation search algorithm accelerated by 3DIV is used to find the optimal rotation to align the two sets of TIVs, which is also the optimal relative rotation between the two raw 3D point sets. Then, the rotationally aligned point sets form the input of the translation search, which is solved by another global optimal algorithm accelerated by 3DIV. The two raw point sets are aligned after both the optimal rotation and the optimal translation are found.}
\label{fig_sim}
\end{figure*}

A TIV can be simply constructed by subtracting a pair of points in the original point set, but not all the TIVs constructed in this way are employed for finding the optimal rotation for three reasons: 1) The total number of TIVs is too large because it is the square of the number of original points. 2) Short TIVs are very sensitive to noise regarding the positions of the original points; consequently, it is difficult to find the optimal rotation by aligning them. 3) The length of a TIV remains unchanged under rotation; therefore, for a TIV $\textbf{m}_{i1-i2}$ from the moving set, a correspondence can be found in only a small subset of all the TIVs from the fixed set that have the same length as $\textbf{m}_{i1-i2}$. By considering all these factors, we choose to match a subset of TIVs with relatively long lengths. Details of the TIVs selection process can be found in Section 6.

\section{Efficient Rotation Search by Utilizing 3D Integral Volume}
After the 6D rigid transformation has been decomposed into one 3D rotation and one 3D translation, these two groups of lower dimensional parameters are searched separately. In this section, we propose a rotation search algorithm to efficiently align the two sets of TIVs constructed from the original point sets. The optimal found rotation is also the 3D rotation portion of the original 6D rigid transformation. We adopt the consensus set maximization-based rotation search explored in [32], [38], [39], and for completeness, we first give the objective function defined on the $L_2$ distance and its geometric bound used in [32] in Sections 4.1 and 4.2, respectively. Then, in Section 4.3, we give the $L_\infty$  distance-based objective function used in this paper to align the two sets of TIVs and the geometric bound used in BnB. In Section 4.4, we propose using the 3DIV data structure to speed up the bound evaluation and present the final rotation search algorithm.

\subsection{Geometric Matching under The $L_2$  Distance}
Let $\mathcal{M}_{\textbf{v}}=\{\textbf{m}_{\textbf{v}i}\}{^M_{i=1}}$ and $\mathcal{S}_{\textbf{v}}=\{\textbf{s}_{\textbf{v}j}\}{^N_{j=1}}$ be the sets of TIVs constructed from the moving and fixed point sets, respectively. Please note that $\mathcal{M}_{\textbf{v}}$ and $\mathcal{S}_{\textbf{v}}$ are both 3D point sets related by an unknown 3D rotation that we want to find by rotation search. $M$ and $N$ are the numbers of TIVs in the moving and fixed point sets, respectively. To align the two sets of TIVs, we want to find a 3D rotation $\textbf{R}$ that maximizes 
\begin{equation}
\mathcal{Q}_{\textbf{r}}(\textbf{R})=\sum\limits_{i}\max\limits_{j} { \lfloor {\left\|{\textbf{Rm}_{\textbf{v}i} - \textbf{s}_{\textbf{v}j}} \right\| \le \epsilon_{\textbf{r}}  }\rfloor },
\end{equation}
where the indicator function $\lfloor \cdot \rfloor $ returns 1 when its predicate $\cdot$ is true and 0 otherwise, $ \left\| \cdot \right\| $ represents the $L_2$ norm, namely, the Euclidean distance between $\textbf{s}_{\textbf{v}j}$ and the rotated $\textbf{m}_{\textbf{v}i}$, and $\epsilon_{\textbf{r}}$ is a predefined inlier threshold. Two 3D points are regarded as a consensus pair when their distance is equal to or less than $\epsilon_{\textbf{r}}$. In geometric matching under the $L_2$ norm, we attempt to find an optimal rotation $\textbf{R}_{opt}$, given $\mathcal{M}_{\textbf{v}}$, $\mathcal{S}_{\textbf{v}}$ and the inlier threshold $\epsilon_{\textbf{r}}$, to maximize $\mathcal{Q}_{\textbf{r}}(\textbf{R})$:
\begin{equation}
\textbf{R}_{opt}=\mathop{\arg\max}_{\textbf{R}\in{SO(3)}}\mathcal{Q}_{\textbf{r}}(\textbf{R};\mathcal{M}_{\textbf{v}},\mathcal{S}_{\textbf{v}},\epsilon_{\textbf{r}}).
\end{equation}

Intuitively, we try to find a rotation that maximizes the number of TIVs in $\mathcal{M}_{\textbf{v}}$  for which we can find a consensus in $\mathcal{S}_{\textbf{v}}$  after being rotated.

\subsection{Representation of Rotation And Geometric Bound}
In this paper, we solve the rotation search problem by utilizing a BnB global optimization framework; therefore, we need to derive an upper and a lower bound of the maximum of $\mathcal{Q}_{\textbf{r}}(\textbf{R})$ in a branch of the parameter space of rotation. A 3D rotation can be parameterized in several different ways, e.g., by quaternions or Euler angles. In this paper, we choose the angle-axis representation, which represents a 3D rotation by a 3D vector whose direction and magnitude represent the axis of the rotation and the radian of the rotation along that axis, respectively. By using the angle-axis representation, all possible 3D rotations lie within a ball of radius $\pi$, which is called the $\pi$-ball. To search for the optimal rotation $\textbf{R}_{opt}$  for (5) within the  $\pi$-ball, we employ the BnB algorithm for global optimality. As in [33], [38], [39], the  $\pi$-ball is embedded in a cube with a side length of 2$\pi$ and then consecutively subdivided into 8 sub-cubes by bisecting each side of the cube (essentially bisecting every dimension of the 3D rotation vector). The BnB algorithm recursively subdivides and prunes the cubes until a termination condition is met, which is commonly defined as occurring when the gap between the upper and lower bound is smaller than a prescribed value. In the BnB search procedure, an upper bound is required for every sub-cube $\mathbb{C}_{\textbf{r}}$  such that
\begin{equation}
\overline{\mathcal{Q}}_{\textbf{r}}(\mathbb{C}_{\textbf{r}})\ge\max\limits_{\textbf{r}\in\mathbb{C}_{\textbf{r}}}\mathcal{Q}_{\textbf{r}}(\textbf{R}_{\textbf{r}}),
\end{equation}
where $\textbf{r}$ is a point in $\mathbb{C}_{\textbf{r}}$  representing a rotation, and $\textbf{R}_{\textbf{r}}$  is the matrix form of $\textbf{r}$. According to [32], in the  $\pi$-ball comprising all the 3D rotation vectors, given a sub-cube $\mathbb{C}_{\textbf{r}}$, an upper bound of (4) can be defined as
\begin{equation}
\overline{\mathcal{Q}}_{\textbf{r}}(\mathbb{C}_{\textbf{r}})=\sum\limits_{i} \max\limits_{j} { \lfloor {\left\|{\textbf{R}_{\textbf{c}}\textbf{m}_{\textbf{v}i} - \textbf{s}_{\textbf{v}j}} \right\| \le \epsilon_{\textbf{r}}+\delta_{\textbf{r}i}  }\rfloor },
\end{equation}
where $\textbf{R}_{\textbf{c}}$  is the matrix form of the rotation vector represented by the midpoint of the diagonal of $\mathbb{C}_{\textbf{r}}$, and $\delta_{\textbf{r}i}$  has the form of
\begin{equation}
\delta_{\textbf{r}i}=\sqrt{2\left\|\textbf{m}_{\textbf{v}i}\right\|^2(1-\text{cos} \alpha_{\mathbb{C}_{\textbf{r}}})},
\end{equation}
where  $\alpha_{\mathbb{C}_{\textbf{r}}}=\left\|\textbf{v}_{\textbf{1}}-\textbf{v}_{\textbf{2}}\right\|/2$, and $\textbf{v}_{\textbf{1}}$, $\textbf{v}_{\textbf{2}}$  denote the two opposite corner points of  $\mathbb{C}_{\textbf{r}}$. Geometrically, by applying the rotation represented by a point $\textbf{R}_{\textbf{r}}$  in the sub-cube  $\mathbb{C}_{\textbf{r}}$, $\textbf{R}_{\textbf{r}}\textbf{m}_{\textbf{v}i}$  lies in a  $\delta_{\textbf{r}i}$-ball whose radius is $\delta_{\textbf{r}i}$  and centered at  $\textbf{R}_{\textbf{c}}\textbf{m}_{\textbf{v}i}$, called the uncertainty ball. An illustrative case is shown in Fig. 3(a). Equation (7) can be regarded as counting the number of uncertainty balls that can find a consensus point in the fixed set. Because the rotated moving point is geometrically constrained within the uncertainty ball, the number of uncertainty balls in which a consensus is found must be equal to or larger than the number of moving points for which a consensus can be found. For more details of the derivation and proof of the upper bound (7), we refer readers to [13], [32], [33], [39].

\subsection{The Geometric Bound under $L_\infty$  Distance}
Following [33], a series of BnB optimization methods were proposed, and many of them attempted to seek a tighter bound [34], [38], [39], [43], which is usually more complex. Nevertheless, the overall runtime of BnB optimization is influenced not only by the tightness of the bound but also by the efficiency with which the bound can be evaluated, as was also noted in [39]. The  $L_2$ norm is used to represent the distance in (4), which results in a  $\delta_{\textbf{r}i}$-ball in the bounding function (7) to form an equidistant surface. When evaluating the bound, extensive calculations of the distances between point pairs are inevitable, forming a major bottleneck when evaluating the upper bound (7). [33] accelerated the bound evaluation by utilizing a distance transform (DT) but at the cost of losing the global optimality [39]. Alternatively, we propose substituting the  $L_2$ norm in (4) with the $L_\infty$  norm; thus, calculating the Euclidean distance is replaced by calculating the Chebyshev distance, leading to our new geometric matching objective function:
\begin{equation}
\mathcal{Q}_{\textbf{r}}(\textbf{R})=\sum\limits_{i} \max\limits_{j} { \lfloor {\left\|{\textbf{Rm}_{\textbf{v}i} - \textbf{s}_{\textbf{v}j}} \right\|_{\infty} \le \epsilon_{\textbf{r}}  }\rfloor }.
\end{equation}

Given a sub-cube  $\mathbb{C}_{\textbf{r}}$ of the rotation space, finding the lower bound of the maximum value of Equation (9) is trivial.

\textbf{Lower bound:} The value of Equation (9) at any point in  $\mathbb{C}_{\textbf{r}}$  can be used as the lower bound in the branch; we choose to use the value at the midpoint of  $\mathbb{C}_{\textbf{r}}$:
\begin{equation}
\underline{\mathcal{Q}}_{\textbf{r}}(\mathbb{C}_{\textbf{r}})=\sum\limits_{i} \max\limits_{j} { \lfloor {\left\|{\textbf{R}_{\textbf{c}}\textbf{m}_{\textbf{v}i} - \textbf{s}_{\textbf{v}j}} \right\|_\infty \le \epsilon_{\textbf{r}}  }\rfloor }
\end{equation}

\textbf{Upper bound:} Similar to the upper bound under the $L_2$  distance in (7), the upper bound under the $L_{\infty}$  distance can be defined as follows:
\begin{equation}
\overline{\mathcal{Q}}_{\textbf{r}}(\mathbb{C}_{\textbf{r}})=\sum\limits_{i} \max\limits_{j} { \lfloor {\left\|{\textbf{R}_{\textbf{c}}\textbf{m}_{\textbf{v}i} - \textbf{s}_{\textbf{v}j}} \right\|_\infty \le \epsilon_{\textbf{r}}+\delta_{\textbf{r}i}  }\rfloor }
\end{equation}

\begin{figure}[htbp]
\centering
\subfloat[]{\includegraphics[width=1.75in]{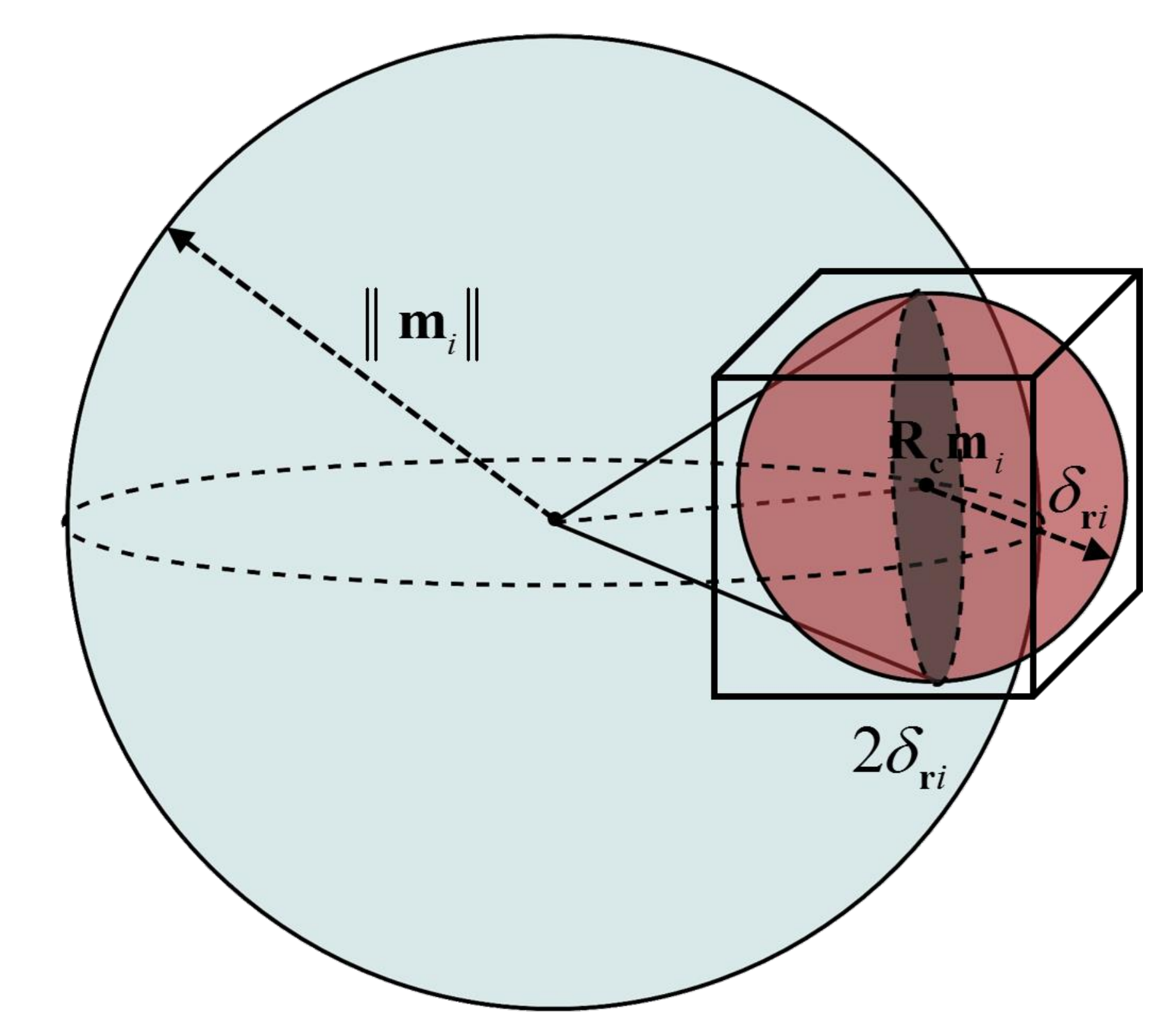}
\label{fig_first_case}}
\hfil
\subfloat[]{\includegraphics[width=1.5in]{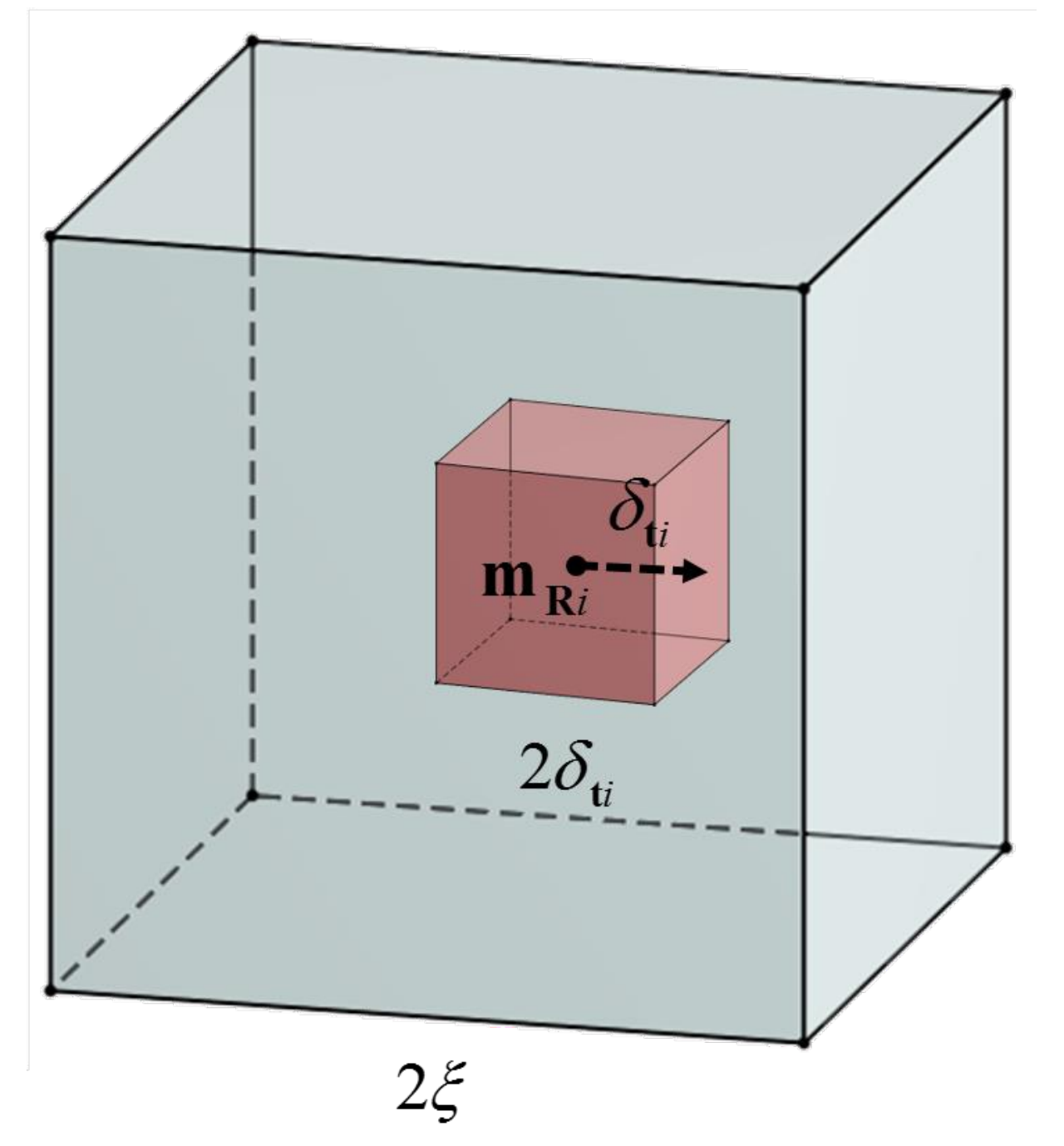}
\label{fig_second_case}}
\caption{An illustration of geometric bounds in the rotation and translation spaces. (a) geometric bounds in the rotation space. The red  $\delta_{\textbf{r}i}$-ball (the uncertainty ball) is from [33] while the gray patch (the uncertainty patch) is from [39]. The cube with a side length of $2\delta_{\textbf{r}i}$  is the cubic bound for the rotation search used in this paper. (b) the geometric bounds in the translation space. The red  $2\delta_{\textbf{r}i}$-cube is the cubic bound for the translation search used in this paper.}
\label{fig_sim}
\end{figure}

Compared to the bounding function (7), we can see that (7) leads to a  $\delta_{\textbf{r}i}$-ball of radius $\delta_{\textbf{r}i}$  centered at  $\textbf{R}_{\textbf{c}}\textbf{m}_{\textbf{v}i}$, while (11) results in a cube with a side length of $2\delta_{\textbf{r}i}$  and also centered at   $\textbf{R}_{\textbf{c}}\textbf{m}_{\textbf{v}i}$ (the cube in Fig. 3(a)). Using our new formulation, seeking a consensus point within a ball is replaced by seeking a consensus point within a cube, which we propose to solve exactly and efficiently by utilizing the 3DIV data structure introduced in the next subsection.

\subsection{Fast Bound Evaluation Based on 3DIV}
In this section, we propose to extend the concept of an integral image, which is a widely used data structure used to rapidly count the sum of elements within a rectangular area, to quickly determine whether a fixed point exists within a cube that forms a consensus pair with a rotated moving point. We denote the 3D version of the integral image as 3D Integral Volume (3DIV). We enclose the fixed point set $\mathcal{P}=\{\textbf{p}_i\}{^P_{i=1}}$  with a cuboid $\mathbb{I}$  that is as small as possible and whose edges are parallel to the coordinate axes. We denote the vertex with the smallest $x$ coordinate and the largest $y$ and $z$ coordinates as the starting vertex. Cuboid  $\mathbb{I}$ is divided into  $n_x$, $n_y$  and $n_z$  subsections along the $x$-,  $y$- and  $z$-axes, respectively, so that there are $(n_x+1)(n_y+1)(n_z+1)$  nodes in  $\mathbb{I}$. As illustrated in Fig. 4, the value of each node is the total number of points in the cuboid that has the line between the node and the starting vertex as its diagonal. This structure of  $(n_x+1)(n_y+1)(n_z+1)$ nodes is the 3DIV structure. We can construct a 3DIV using the following method: for each point $\textbf{p}_i$ in  $\mathcal{P}$, we add 1 to every node whose $x$  coordinate is larger than the $x$ coordinate of  $\textbf{p}_i$ but whose $y$  and $z$  coordinates are smaller than the $y$  and $z$  coordinates of  $\textbf{p}_i$. We present the 3DIV construction algorithm in \textbf{Algorithm 1}.

Given the 3DIV, the number of points in any cuboid can be rapidly calculated from the values of its eight nodes. As illustrated in Fig. 5, we first calculate the number of points in the red cuboid in Fig. 5(a). Based on the 3DIV construction method, the value of node A is the number of points in parts I and IV; the value of node B is the number of points in parts I, II, III and IV; the value of node C is the number of points in parts III and IV, and the value of node D is the number of points in part IV. Then, we can obtain the number of points in part II by the following simple calculation:
\begin{equation}
\begin{aligned}
&\text{value}(\text{B})-\text{value}(\text{A})-\text{value}(\text{C})+\text{value}(\text{D})\\
=&(N_I+N_{II}+N_{III}+N_{IV})-(N_I+N_{IV})-\\
&(N_{III}+N_{IV})+N_{IV}\\
=&N_{II}
\end{aligned}
\end{equation}

\begin{figure}[htbp]
\centering
\includegraphics[width=2in]{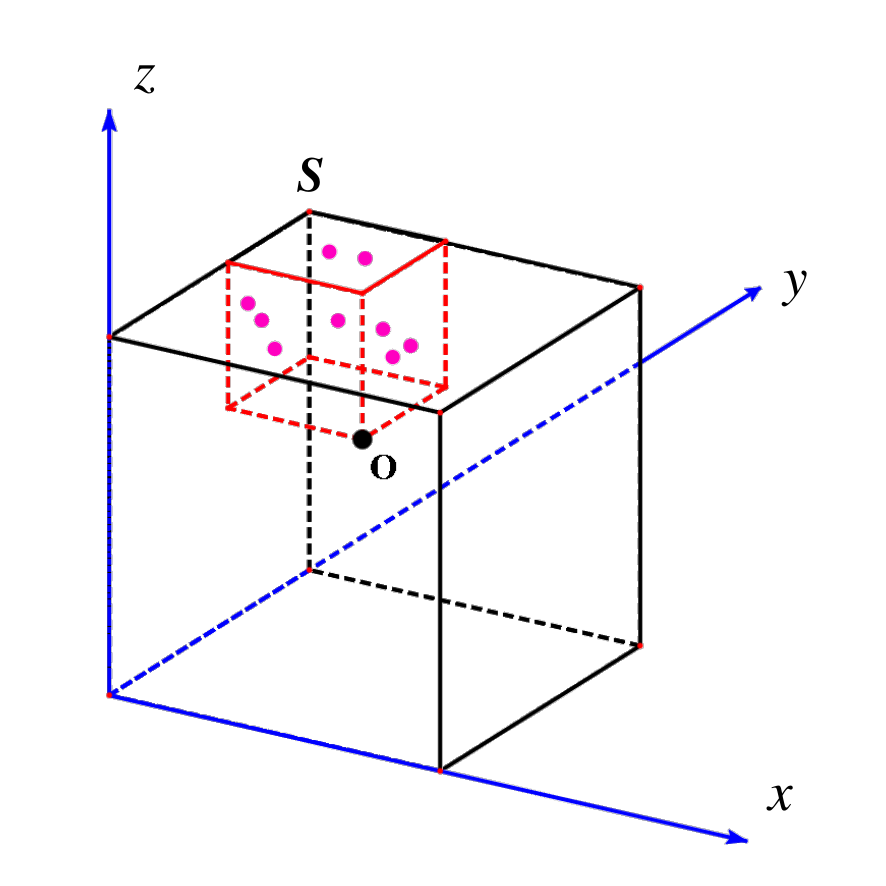}
\caption{An illustration of 3DIV and its construction. $\textbf{S}$ is the starting vertex of the 3DIV. There are 9 points in the small red cube so the value of O is 9.}
\label{fig_sim}
\end{figure}

\begin{algorithm}
\caption{3D Integral Volume Construction}
\label{alg1}
\begin{algorithmic}[1]
\REQUIRE A 3D point set $\mathcal{P}=\{ \textbf{p}_i \}{^P_{i=1}} $, the number of subsections $n$ along every dimension. 
\STATE Divide $x$-axis, $y$-axis and $z$-axis into $n$ subsections to get $\text{index}\_x=\{x_e\}{^n_{e=1}}$, $\text{index}\_y=\{y_f\}{^n_{f=1}}$ and $\text{index}\_z=\{z_g\}{^n_{g=1}}$, $(n+1)^3$ nodes $N(x,y,x)$ with $V(N)=0$.
\FOR {${\textbf{p}_i}\in{\mathcal{P}}$}
\STATE find the position of $x_{\textbf{p}_i}$ in  $\text{index}\_x$: $x_e\le{x_{\textbf{p}_i}}<x_{e+1}$.
\STATE find the position of $y_{\textbf{p}_i}$ in  $\text{index}\_y$: $y_f\le{y_{\textbf{p}_i}}<y_{f+1}$.
\STATE find the position of $z_{\textbf{p}_i}$ in  $\text{index}\_z$: $z_g\le{z_{\textbf{p}_i}}<z_{g+1}$.
\FOR{node $N(x,y,z)$ whose $x\ge{x_e}, y\le{y_f}, z\le{z_g}$}
         \STATE $V(N)=V(N)+1$
         \ENDFOR
\ENDFOR

\RETURN the 3D integral Volume $\mathbb{I}$ with $(n+1)^3$ nodes.
\end{algorithmic}
\end{algorithm}

Therefore, we know the number of points in part II, which is a cuboid with ABCD as its bottom surface. Actually, we can calculate the number of points in any cuboid whose upper surface lies on the upper surface of the whole cuboid  $\mathbb{I}$. Then, as shown in Fig. 5(b), we can compute the number of points lying in the light green cuboid with abcd as its bottom in the same way. By subtracting the number of points in part II and the number of points in the light green cuboid in Fig. 5(b), we can obtain the number of points in the cuboid with ABCD and abcd as its vertices.

The BnB algorithm utilizing the upper bound (11) and the fast bound evaluation method based on 3DIV is presented in \textbf{Algorithm 2}. After initializing the 3DIV, which is very fast, it is both simple and fast to determine whether a scene point exists within an arbitrary cuboid that forms a consensus pair. When the vertices of the uncertainty cube do not fall on the nodes of the 3DIV, we can use a slightly larger cuboid to check the consensus. Geometrically, it is obvious that the cubic bound used in this paper is looser than the ball bound used in [32], not to mention the uncertainty patch used in [39]. However, we would such as to emphasize again that due to its simple calculation of addition and subtraction of 8 values, using our cubic bound can realize a very fast rotation search on TIVs, which will be demonstrated in Section 6.

\begin{figure}[tbp]
\centering
\subfloat[]{\includegraphics[width=1.7in]{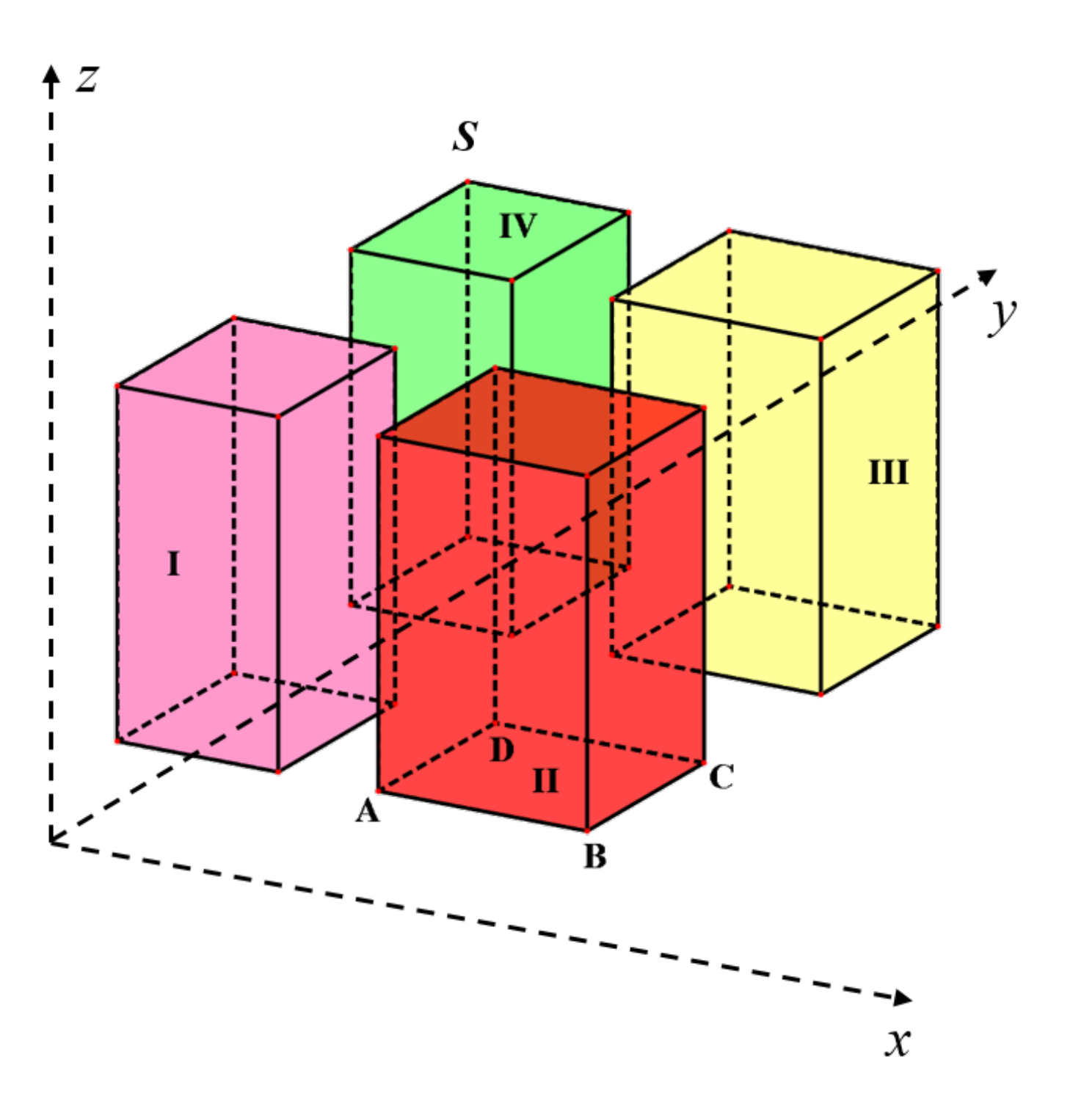}
\label{fig_first_case}}
\hfil
\subfloat[]{\includegraphics[width=1.7in]{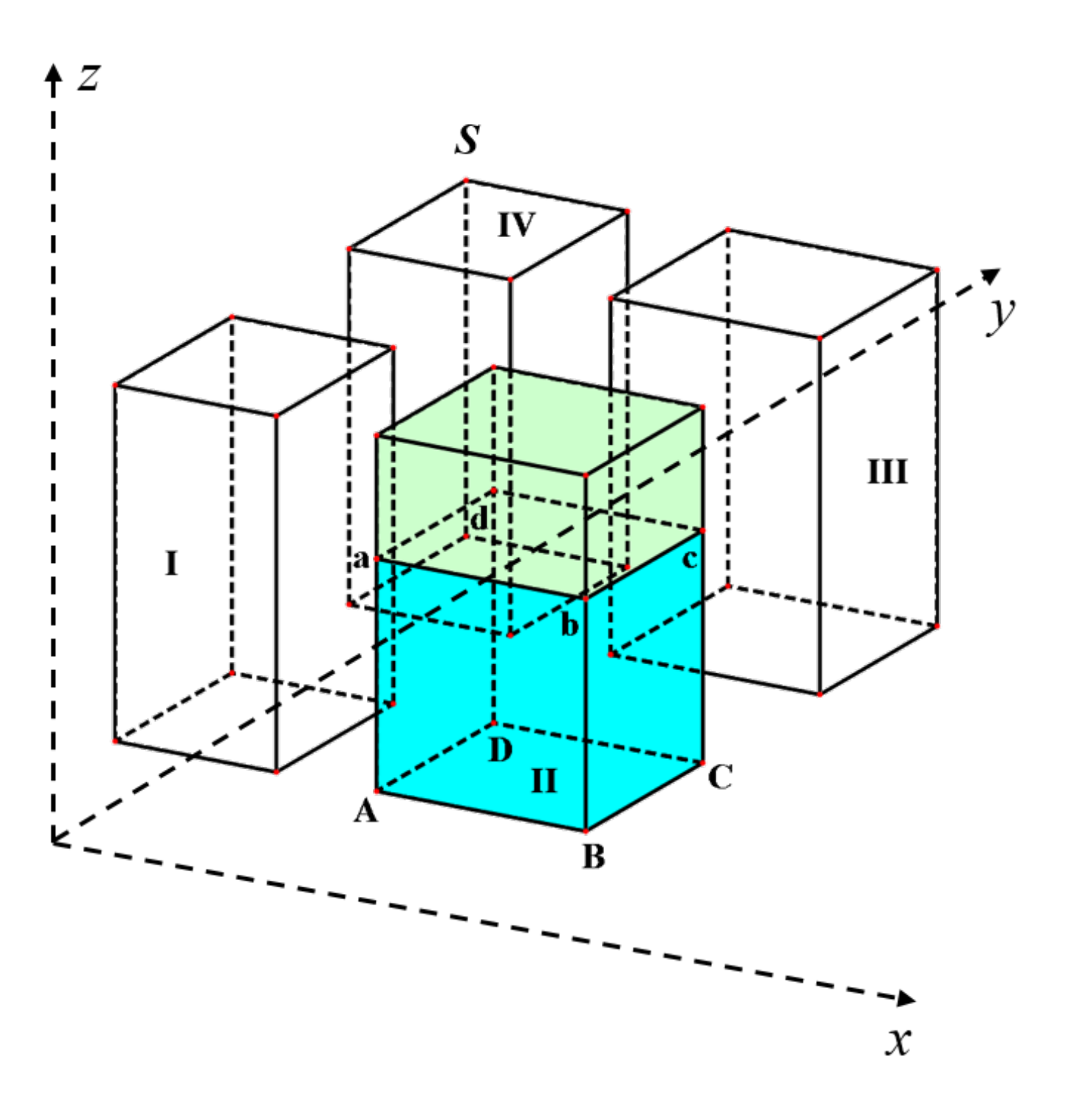}
\label{fig_second_case}}
\caption{Illustration of how to calculate the number of points within a cuboid using 3DIV. We want to calculate the number of points in the light blue cuboid in (b), which has vertices at A, B, C and D on its lower surface and at a, b, c and d on its upper surface. As illustrated in (a), we divide the cuboid with B and the starting vertex as its diagonal points into four parts and denote them as parts I, II, III and IV, which are drawn separately for better illustration.}
\label{fig_sim}
\end{figure}

\begin{algorithm}
\caption{BnB Rotation Search to Maximise (9)}
\label{alg1}
\begin{algorithmic}[1]
\REQUIRE $\mathcal{M}_{\textbf{v}}=\{ \textbf{m}_{\textbf{v}i} \}{^M_{i=1}} $, the 3D Integral Volume $\mathbb{I}$, inlier threshold $\epsilon_{\textbf{r}}$, the $gap$ between the upper bound $\overline{\mathcal{Q}}_{\textbf{r}}$ and the lower bound $\mathcal{Q}_{\textbf{r}max}$

\STATE Initialise priority queue $q$, $\mathcal{Q}_{\textbf{r}max}\leftarrow 0$, $\textbf{R}_{opt}\leftarrow \phi$.\\
Insert $\mathbb{C}_{\textbf{r}}$ with priority $\overline{\mathcal{Q}}_{\textbf{r}}(\mathbb{C}_{\textbf{r}})$ into $q$.

\WHILE{$q$ is not empty}
\STATE Obtain the highest priority cube $\mathbb{C}_{\textbf{r}}$ from $q$.
\IF{$\overline{\mathcal{Q}}_{\textbf{r}}(\mathbb{C}_{\textbf{r}})-\mathcal{Q}_{\textbf{r}max}<gap$}
\STATE Terminate.
\ENDIF
\STATE $\textbf{R}_{\textbf{c}}\leftarrow$ center point of $\mathbb{C}_{\textbf{r}}$.
\IF{$\mathcal{Q}_{\textbf{r}}(\textbf{R}_{\textbf{c}})>\mathcal{Q}_{\textbf{r}max}$}
\STATE $\mathcal{Q}_{\textbf{r}max}\leftarrow \mathcal{Q}_{\textbf{r}}(\textbf{R}_{\textbf{c}}), \textbf{R}_{opt}\leftarrow \textbf{R}_{\textbf{c}}$.
\ENDIF
\STATE Subdivide $\mathbb{C}_{\textbf{r}}$ into eight cubes $\{\mathbb{C}_{\textbf{r}d}\}{^8_{d=1}}$.
\FOR{each $\mathbb{C}_{\textbf{r}d}$}
\IF{$\overline{\mathcal{Q}}_{\textbf{r}}(\mathbb{C}_{\textbf{r}d})>\mathcal{Q}_{\textbf{r}max}$}
\STATE Insert $\mathbb{C}_{\textbf{r}d}$ with priority $\overline{\mathcal{Q}}_{\textbf{r}}(\mathbb{C}_{\textbf{r}d})$ into $q$.
\ELSE 
\STATE Discard $\mathbb{C}_{\textbf{r}d}$
\ENDIF
\ENDFOR
\ENDWHILE
\RETURN Optimal rotation $\textbf{R}_{opt}$ with quality $\mathcal{Q}_{\textbf{r}max}$.
\end{algorithmic}
\end{algorithm}

\section{Efficient Translation Search}
By using the above global rotation search on TIVs, the optimal rotation $\textbf{R}_{opt}$  between the original moving and fixed point sets can be found. The original moving point set $\mathcal{M}=\{\textbf{m}_i\}{^M_{i=1}}$  can be rotated by  $\textbf{R}_{opt}$; we denote the rotated point set as  $\mathcal{M}_{\textbf{R}}=\{\textbf{m}_{\textbf{R}i}\}{^M_{i=1}}$. The next step is to find the optimal translation between $\mathcal{M}_{\textbf{R}}$ and $\mathcal{S}$, which can also be solved globally in the BnB framework.

The parameterization of translation is straightforward. When we use a 3D vector $t=[\alpha,\beta,\gamma]$ to represent translation, we can set a range for each coordinate, which results in a cuboid initial translation branch. In practice, we use a fixed range that is large enough for all three coordinates; therefore, the initial translation searching branch is a cube. All the sub-branches in the translation search are cubic.

Similar to (9), we can define the geometric matching objective function for the translation search as follows:
\begin{equation}
\mathcal{Q}_{\textbf{t}}(\textbf{t})=\sum\limits_{i} \max\limits_{j} { \lfloor {\left\|{\textbf{m}_{\textbf{R}i} + \textbf{t}- \textbf{s}_{j}} \right\| _{\infty} \le \epsilon_{\textbf{t}}  }\rfloor } 
\end{equation}

Then, the lower bound and the upper bound of the maximum of function (13) for a translation branch $\mathbb{C}_{\textbf{t}}$  are, respectively, 
\begin{equation}
\underline{\mathcal{Q}}_{\textbf{t}}(\mathbb{C}_{\textbf{t}})=\sum\limits_{i} \max\limits_{j} { \lfloor {\left\|{\textbf{m}_{\textbf{R}i} +\textbf{t}_{\textbf{c}}- \textbf{s}_{j}} \right\| _{\infty} \le \epsilon_{\textbf{t}}  }\rfloor }
\end{equation}
\begin{equation}
\overline{\mathcal{Q}}_{\textbf{t}}(\mathbb{C}_{\textbf{t}})=\sum\limits_{i} \max\limits_{j} { \lfloor {\left\|{\textbf{m}_{\textbf{R}i} +\textbf{t}_{\textbf{c}}- \textbf{s}_{j}} \right\| _{\infty} \le \epsilon_{\textbf{t}}} + \delta_{\textbf{t}i}\rfloor } 
\end{equation}
where $\epsilon_{\textbf{t}}$  is the predefined inlier threshold,  $\textbf{t}_{\textbf{c}}$ is the midpoint of sub-cube $\mathbb{C}_{\textbf{t}}$  and  $\delta_{\textbf{t}i}$ is half the edge length of sub-cube  $\mathbb{C}_{\textbf{t}}$. Note that we use the same inlier threshold for both the rotation search objective function (9) and the translation search objective function (13). Fig. 3(b) shows an illustration of the translation space and the geometric bounds. To rapidly evaluate the upper bound (15), we utilize 3DIV again, but this time it is built from the scene set  $\mathcal{S}=\{\textbf{s}_j\}{^N_{j=1}}$. The BnB algorithm for the translation search is presented in \textbf{Algorithm 3}.

\begin{algorithm}
\caption{BnB Translation Search to Maximise (13)}
\label{alg1}
\begin{algorithmic}[1]
\REQUIRE $\mathcal{M}=\{ \textbf{m}_{i} \}{^M_{i=1}} $, the 3D Integral Volume $\mathbb{I}$, inlier threshold $\epsilon_{\textbf{t}}$, the $gap$ between the upper bound $\overline{\mathcal{Q}}_{\textbf{t}}$ and the lower bound $\mathcal{Q}_{\textbf{t}max}$. 

\STATE Initialise priority queue $q$, $\mathcal{Q}_{\textbf{t}max}\leftarrow 0$, $\textbf{t}_{opt}\leftarrow \phi$.\\
Insert $\mathbb{C}_{\textbf{t}}$ with priority $\overline{\mathcal{Q}}_{\textbf{t}}(\mathbb{C}_{\textbf{t}})$ into $q$.

\WHILE{$q$ is not empty}
\STATE Obtain the highest priority cube $\mathbb{C}_{\textbf{t}}$ from $q$.
\IF{$\overline{\mathcal{Q}}_{\textbf{t}}(\mathbb{C}_{\textbf{t}})-\mathcal{Q}_{\textbf{t}max}<gap$}
\STATE Terminate.
\ENDIF
\STATE $\textbf{t}_{\textbf{c}}\leftarrow$ center point of $\mathbb{C}_{\textbf{t}}$.
\IF{$\mathcal{Q}_{\textbf{t}}(\textbf{t}_{\textbf{c}})>\mathcal{Q}_{\textbf{t}max}$}
\STATE $\mathcal{Q}_{\textbf{t}max}\leftarrow \mathcal{Q}_{\textbf{t}}(\textbf{t}_{\textbf{c}}), \textbf{t}_{opt}\leftarrow \textbf{t}_{\textbf{c}}$.
\ENDIF
\STATE Subdivide $\mathbb{C}_{\textbf{t}}$ into eight cubes $\{\mathbb{C}_{\textbf{t}d}\}{^8_{d=1}}$.
\FOR{each $\mathbb{C}_{\textbf{t}d}$}
\IF{$\overline{\mathcal{Q}}_{\textbf{t}}(\mathbb{C}_{\textbf{t}d})>\mathcal{Q}_{\textbf{t}max}$}
\STATE Insert $\mathbb{C}_{\textbf{t}d}$ with priority $\overline{\mathcal{Q}}_{\textbf{t}}(\mathbb{C}_{\textbf{t}d})$ into $q$.
\ELSE
\STATE Discard $\mathbb{C}_{\textbf{t}d}$
\ENDIF
\ENDFOR
\ENDWHILE
\RETURN Optimal translation $\textbf{t}_{opt}$ with quality $\mathcal{Q}_{\textbf{t}max}$. 
\end{algorithmic}
\end{algorithm}

\section{Experiments}
In this section, we evaluate the performance of the proposed rigid point set registration method and compare it with state-of-the-art global methods using both synthetic and real data.

We denote the proposed 3D rigid point set registration method as TIV-3DIV, which consists of a rotation search on TIVs (\textbf{Algorithm 2}) followed by a translation search (\textbf{Algorithm 3}), and both the rotation search and the translation search are accelerated by 3DIV structures constructed using \textbf{Algorithm 1}. We compared TIV-3DIV to the following global point set registration methods: Go-ICP [33], Glob-GM-ML [39] and APM [35]. GOGMA [34] and [36] are not included as comparison methods because they both take advantage of a GPU.

In addition, by constructing the TIVs, we decompose the 6D space search of rigid transformation into a 3D searches of rotation and translation spaces. The rotation search and the translation search are performed separately; consequently, any rotation search kernel can be used as a substitute for the rotation search algorithm used in TIV-3DIV to form a new rigid registration method, and such new methods can also take advantage of the efficiency of our transformation decomposition framework. To emphasize the significance of the transformation decomposition framework, we implemented two new algorithms by substituting existing rotation kernels for the rotation search part of TIV-3DIV. The two rotation kernels used to create the new algorithms are MCIRC-ML and MCIRC, which were proposed in [38], [39] and are the state-of-the-art methods for rotation search. The two kernels are identical except that matchlist, which is a technique to speed up the algorithm, is removed in MCIRC. The two new methods for 6D rigid registration are denoted as TIV-MCIRC-ML and TIV-MCIRC; these methods execute rotation searches on TIVs using MCIRC-ML and MCIRC, respectively; however, the translation search is performed in the same way as in TIV-3DIV. These two new algorithms were added to the comparison list and help to obtain a fairer comparison between our rotation kernel and MCIRC to show the influence of the tightness and efficiency of bounds on the runtime.

The source code for the compared methods were released by the authors. Go-ICP and Glob-GM-ML are implemented in C++, while APM is implemented using MATLAB. Our algorithms are all implemented in C. All the experiments were conducted on a computer equipped with an Intel Xeon E5 3.2 GHz CPU.

\subsection{Synthetic Data}
The synthetic data used in this experiment include 11 models from the four datasets presented in Fig. 6: Armadillo, Bunny, Dragon, Happy Buddha, Asian Dragon and Statuette from the Stanford 3D Scanning Repository [50], Chicken, Rhino and T-rex from Mian’s dataset [51], [52], Camera from the Stefan Hinterstoisser’s dataset [53] and Hand from the Large Geometric Models Archive at Georgia Tech [54]. The number of points contained in these models ranges from 18,995 to 4,999,996 and are randomly downsampled to a manageable scale in all the experiments. Compared to extracting key points for registration, downsampling is a much simpler preprocessing step and is more widely applicable.

We compare the performance of each method under three common types of degradation of the point sets: inserting some nondata points to one set (outliers), deleting some data points from one set (missing points), and perturbing the position of points (noise) in one set. All the models are randomly downsampled to 500 points, which is sufficient to preserve the 3D shapes. The point sets were uniformly scaled to fit in a cube  $[0,1]^3$, and the translation search range was set to  $[-1,1]^3$, which is large enough to find the global optimal translation.
\begin{figure}[tbp]
\centering
\includegraphics[width=3.4in]{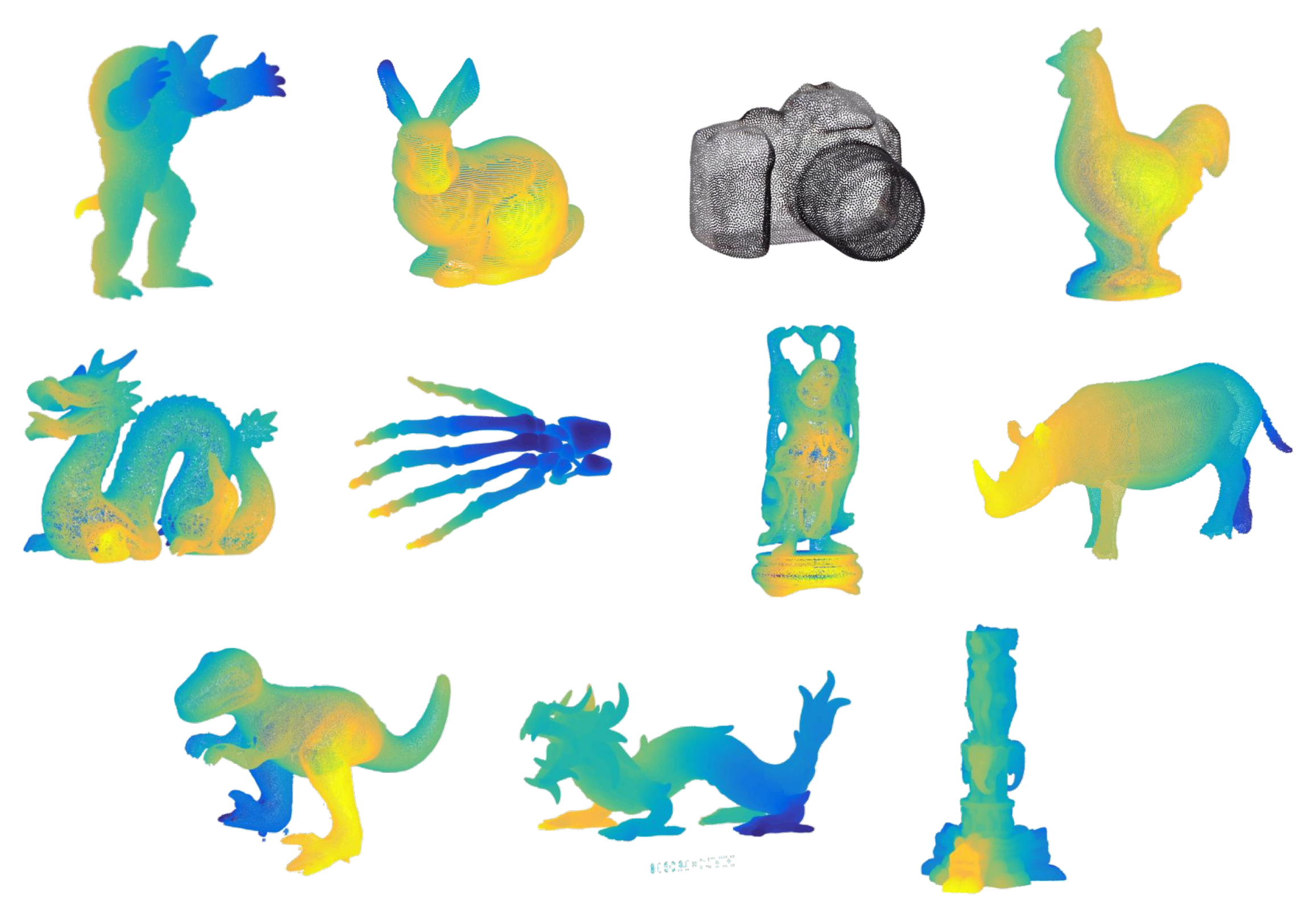}
\caption{The models used in the experiments. From left top: Armadillo, Bunny, Camera, Chicken, Dragon, Hand, Happy Buddha, Rhino, T-rex, Asian dragon and Statuette. }
\end{figure}

To evaluate the registration performance, we employed four metrics. 1) Runtime. This metric included the time needed by every part of the algorithm. For TIV-3DIV, it included the time for constructing TIVs, and for Go-ICP, it included the time for constructing the DT. 2) Angular error. This metric is the error of the found optimal rotation with respect to the ground truth rotation. 3). Translation error. This metric is the error of the found optimal translation with respect to the ground truth translation. 4). RMS. This metric is the root mean square distance between the true corresponding points when the floating point set is transformed by the found optimal rotation and translation. Note that APM is used to calculate the optimal affine matrix instead of the rotation matrix and translation vector, so for APM we do not calculate the angular error. To avoid excessive runtimes, we set a limit for every method: if it was unable to converge and return a final solution within 1,000 s, we terminate it by force.

The algorithm-specific parameters are as follows: for Go-ICP, the trimming percentage is zero, the resolution of DT is  $300\times300\times300$, and the convergence threshold is 0.001; thus, we denote it as Go-ICP (0.001). For APM, we tested two implementations with different distance error $\epsilon_{d}$  choices, 0.1 and 0.3, and denote them as APM (0.1) and APM (0.3), respectively; there is no regularization for APM. The inlier thresholds of the rotation search kernel of TIV-3DIV, TIV-MCIRC-ML, TIV-MCIRC and Glob-GM-ML are the same, which ensures that the red ball will has the same radius as shown in Fig. 3(a). For all the experiments, the resolution of 3DIV is set to  $51\times51\times51$.

\textbf{Inserted nondata points.} For each model, the down-sampled point set was used as the model set; then, we added different ratios of nondata points and applied random rotations and translations to the model set to generate the scene set. For TIV-3DIV, TIV-MCIRC-ML and TIV-MCIRC, the TIVs were selected according to the following strategy: construct all the TIVs and then delete the 5,000 TIVs with the largest norms. Then, pick 200 TIVs with the largest norms from the remaining set. Because of outliers (the nondata points added in this experiment), the TIVs with the largest norms have a high probability of being outliers and are not suitable for rotation search; therefore, they were deleted. The inlier threshold was set as 0.005 for TIV-3DIV, TIV-MCIRC-ML, TIV-MCIRC and Glob-GM-ML.

\textbf{Deletion of some data points.} Different ratios of data points were deleted from the down-sampled point sets to generate the model sets, while the scene sets were constructed by applying random rotations and translations to the entire down-sampled point set. The TIV selection strategy and inlier threshold were the same as those used in the experiment where nondata points were inserted.

\textbf{Perturbations of data points.} For each model, the down-sampled point set was used as the model set, and Gaussian noise with different standard deviations and random rotations and translations was applied to the model set to generate the scene set. In this situation, we selected the 200 TIVs with the largest norms. In this experiment, the inlier threshold of TIV-based methods was set to 0.01, which is the largest standard deviation of Gaussian noise used in the experiment.

The average runtimes of 20 registrations for each setting of 6D rigid registrations for each model are presented in Fig. 7, Fig. 8 and Fig. 9 for the above three experimental situations, respectively. Note that APM (0.1), APM (0.3) and Glob-GM-ML are not shown in the figures because APM (0.1) and Glob-GM-ML were unable to converge and terminate within 1,000 s in all experiments. Although the runtime of APM (0.3) was below 1,000 s in some experiments, its runtime was much longer than those of the other methods. Therefore, it is not plotted in the figures to improve the runtime illustration of the other methods. The results can be analyzed from the following three perspectives:

\begin{figure}[bp]
\centering
\includegraphics[width=3.5in]{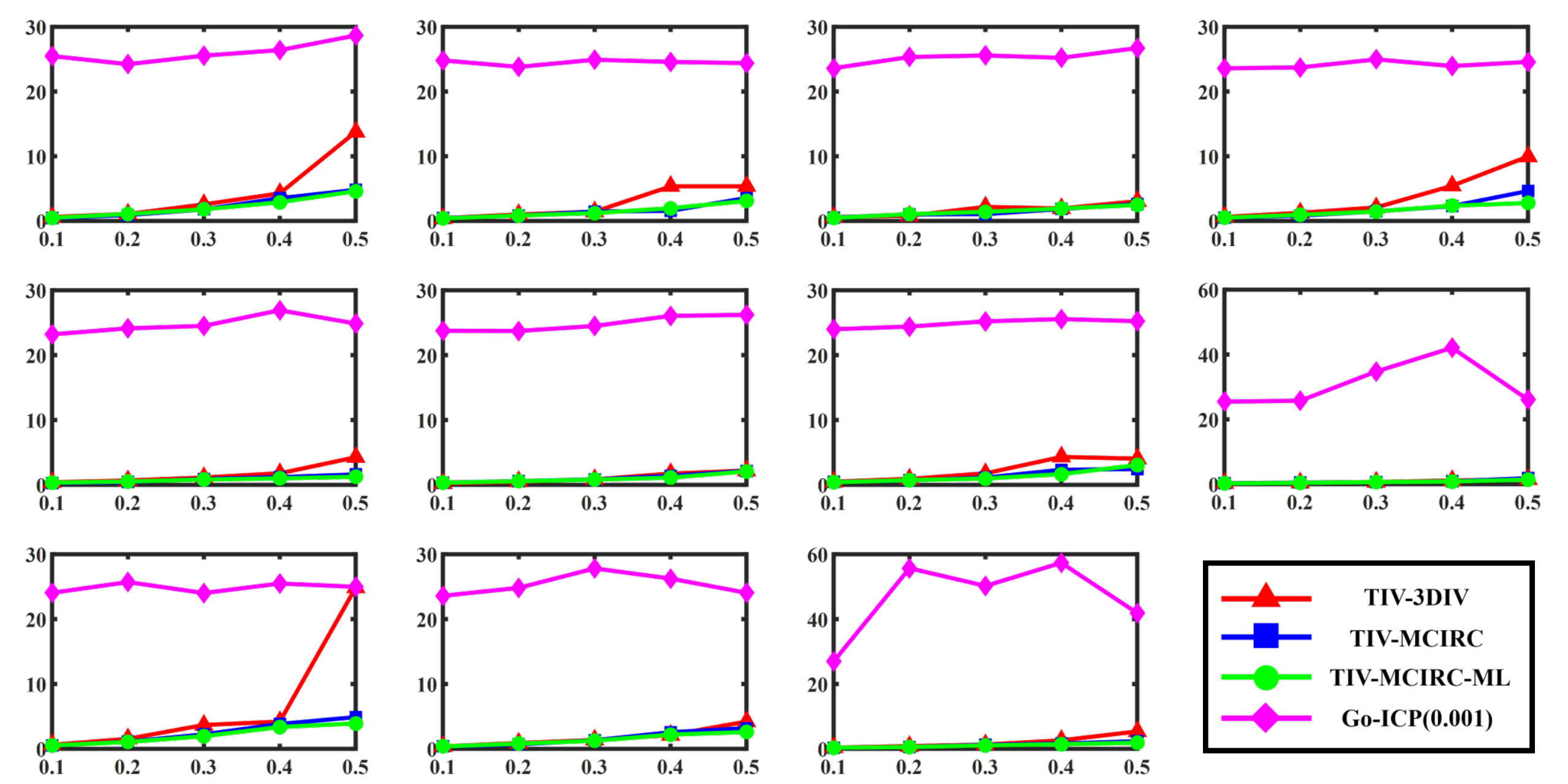}
\caption{Average runtime of 20 runs of 6DOF rigid registration with respect to the outlier ratio for each model in the experiments on insertion of nondata points. The horizontal axis is the outlier fraction, and the vertical axis is the runtime (in seconds). From left top: Armadillo, Bunny, Camera, Chicken, Dragon, Hand, Happy Buddha, Rhino, T-rex, Asian dragon and Statuette. }
\end{figure}
 
\begin{figure}[htbp]
\centering
\includegraphics[width=3.6in]{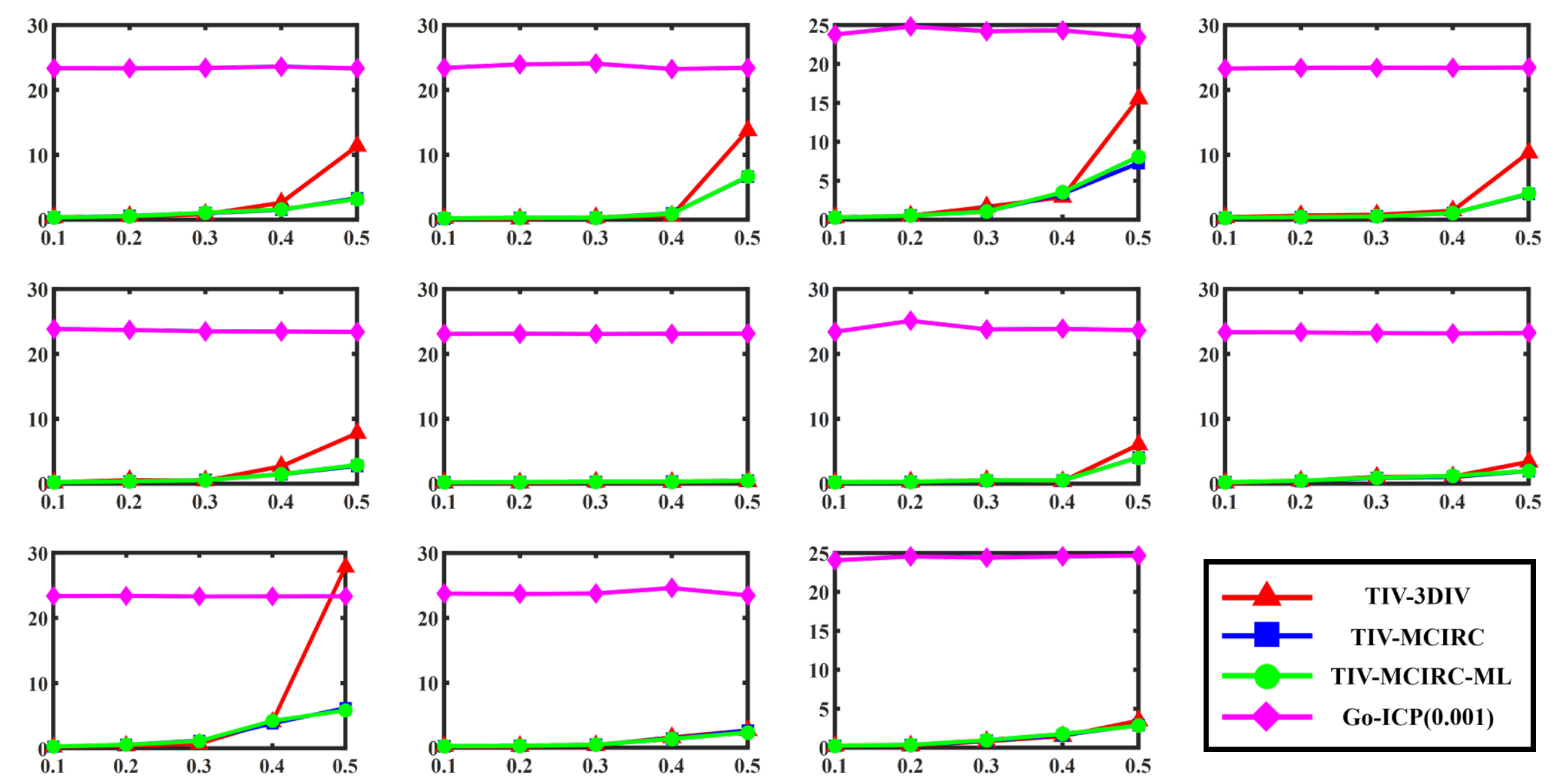}
\caption{Average runtime of 20 runs of 6DOF rigid registration with respect to the missing point fraction for each model in the experiment where some data points were deleted. The horizontal axis is the missing point fraction, and the vertical axis is the runtime (in seconds). From left top: Armadillo, Bunny, Camera, Chicken, Dragon, Hand, Happy Buddha, Rhino, T-rex, Asian dragon and Statuette.}
\end{figure}

\begin{figure}[htbp]
\centering
\includegraphics[width=3.6in]{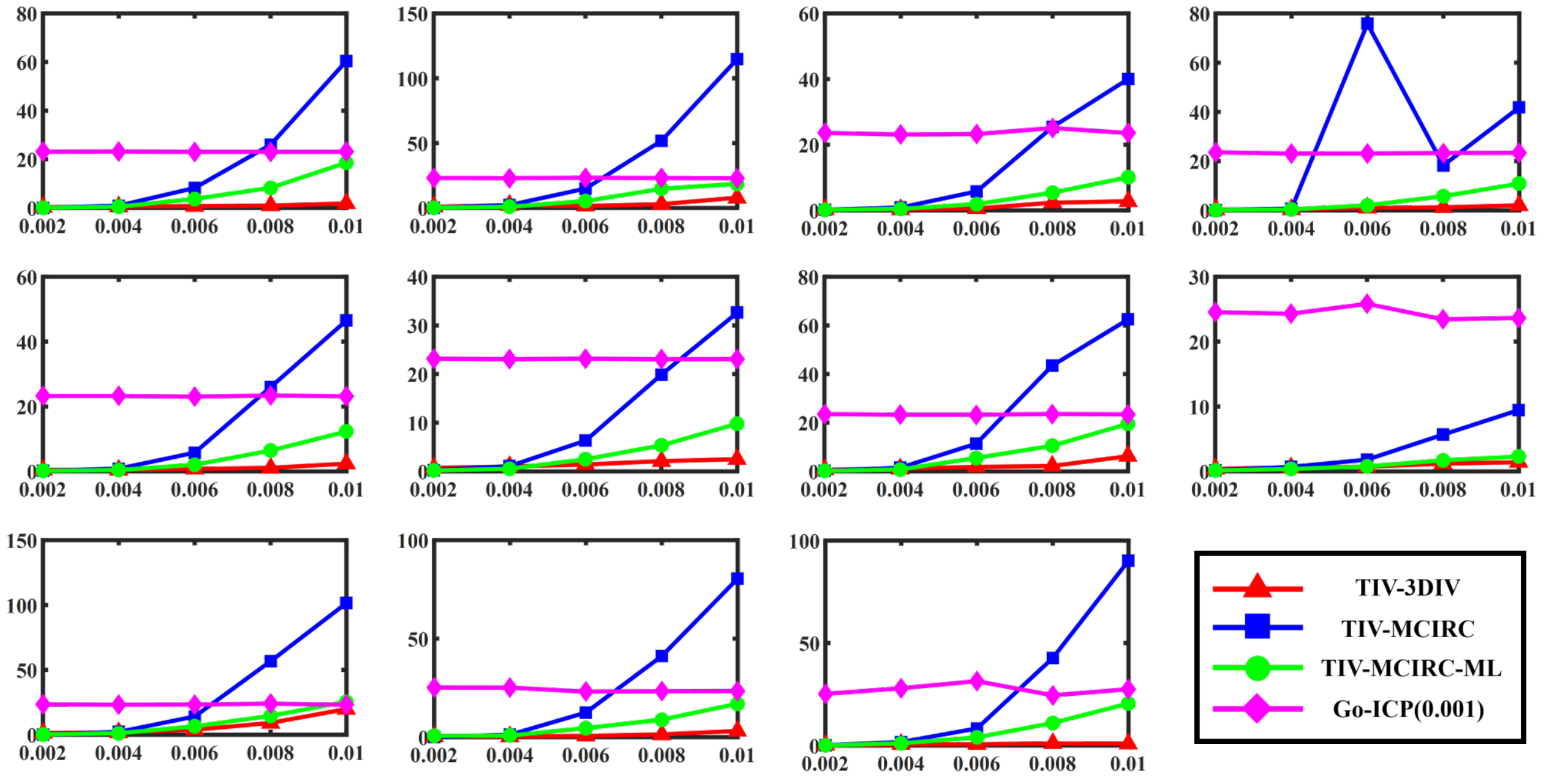}
\caption{Average runtime of 20 runs of 6DOF rigid registration with respect to the noise level for each model in the experiments with perturbations of data points. The horizontal axis is the standard deviation of the Gaussian noise added to the data point’s coordinates, and the vertical axis is the runtime (in seconds). From left top: Armadillo, Bunny, Camera, Chicken, Dragon, Hand, Happy Buddha, Rhino, T-rex, Asian dragon and Statuette. }
\end{figure}

\textbf{1). The comparison between Glob-GM-ML and TIV-MCIRC-ML, TIV-MCIRC, demonstrates the superiority of our proposed TIV-based framework of transformation decomposition.} Considering that the three methods have the same rotation kernel (except that the matchlist was turned off in TIV-MCIRC), the runtime comparison clearly demonstrates the superiority of our TIV-based framework of transformation decomposition, which breaks a 6D BnB search problem into two 3D BnB search problems. Glob-GM-ML was unable to converge in the allotted 1,000 s in all the experiments$-$in fact, its runtime exceeded 3,000 s for most of the experiments. However, when the same rotation kernel was embedded in the TIV-based decomposition framework, the runtime was reduced by three orders of magnitude, especially in the first two situations. The runtimes of TIV-MCIRC and TIV-MCIRC-ML are similar in the first two situations, but TIV-MCIRC-ML was faster in the last situation, in which there were perturbations of the data points. This situation seemed to be more difficult than the first two situations for both TIV-MCIRC-ML and TIV-MCIRC.

\textbf{2). The comparison between TIV-3DIV and Go-ICP, APM, Glob-GM-ML shows that the proposed TIV-3DIV is much faster than the state-of-the-art methods.} This experiment involves a direct comparison among the proposed method and the state-of-the-art global methods for 6D rigid point set registration. First, TIV-3DIV is three orders of magnitude faster than APM and Glob-GM-ML and approximately ten times faster than Go-ICP accelerated by DT. From the previous comparison, we can see that the rotation kernel of Glob-GM-ML was efficient. Nevertheless, Glob-GM-ML was very slow. The primary reason is that Glob-GM-ML searches over a 6D parameter space, and the computational complexity of BnB is exponential to the problem dimensionality. APM actually used an efficient and tight bound, but it optimized over the parameters of the affine matrix, whose dimensionality is even higher. Although Go-ICP utilizes a nested strategy to search for translation and rotation, similar to Glob-GM-ML, and its bound is even looser than that of Glob-GM-ML, it is much faster than Glob-GM-ML. The distance transform technique, which is used to efficiently find the approximated distance of a point to its nearest neighbor, contributed most to the efficiency of Go-ICP. TIV-3DIV achieved the best runtime. Although the angular error or translation error of TIV-3DIV was not the smallest, its accuracy was good enough considering it is a global method, and its accuracy can easily be improved with local refinement.

\textbf{3). The comparison between TIV-3DIV and TIV-MCIRC-ML, TIV-MCIRC.} These three methods share the same transformation decomposing framework and the same translation search algorithm; thus, the comparison results reveal the differences of their rotation kernels. In the first two situations, the three resulted in a near-tie, while in the situation with data point perturbations, TIV-3DIV substantially outperformed the other two. To provide a better demonstration without the influence of the translation search, we conducted a pure rotation search on the TIVs. The experimental setting remained the same as in the 6D experiments, except that we extracted the rotation kernel and only performed the rotation search on TIVs. The average runtimes are shown in Fig. 10, Fig. 11 and Fig. 12, where the proposed rotation search approach based on 3DIV is denoted by 3DIV-RS. The results are similar to those in the 6D experiments: both types of experiments show that in our TIV-based framework, the proposed 3DIV-based rotation search is at least as fast as MCIRC$\backslash$MCIRC-ML, and it is faster when perturbations exist on the points’ positions. These results support our hypothesis that the BnB algorithm with a looser but rapidly evaluable bound runs faster than when a tighter but complex and slow-to-evaluate bound is used.

\begin{figure}[bp]
\centering
\includegraphics[width=3.6in]{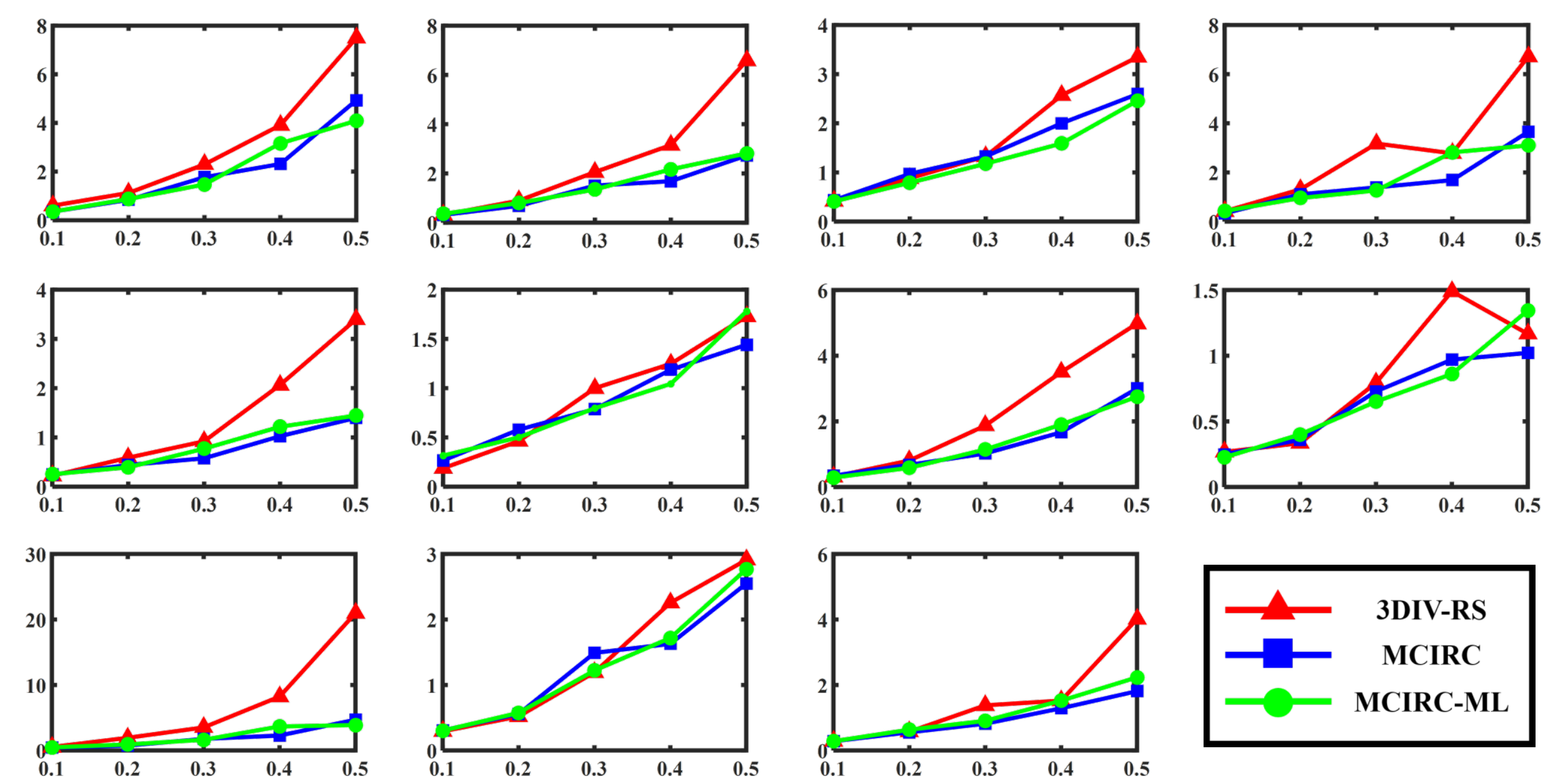}
\caption{Average runtime of 20 runs of 3DOF rotation search with respect to the outlier fraction for each model in the experiments with insertion of nondata points. The horizontal axis is the outlier fraction, and the vertical axis is the runtime (in seconds). From left top: Armadillo, Bunny, Camera, Chicken, Dragon, Hand, Happy Buddha, Rhino, T-rex, Asian dragon and Statuette. }
\end{figure}

\begin{figure}[tbp]
\centering
\includegraphics[width=3.6in]{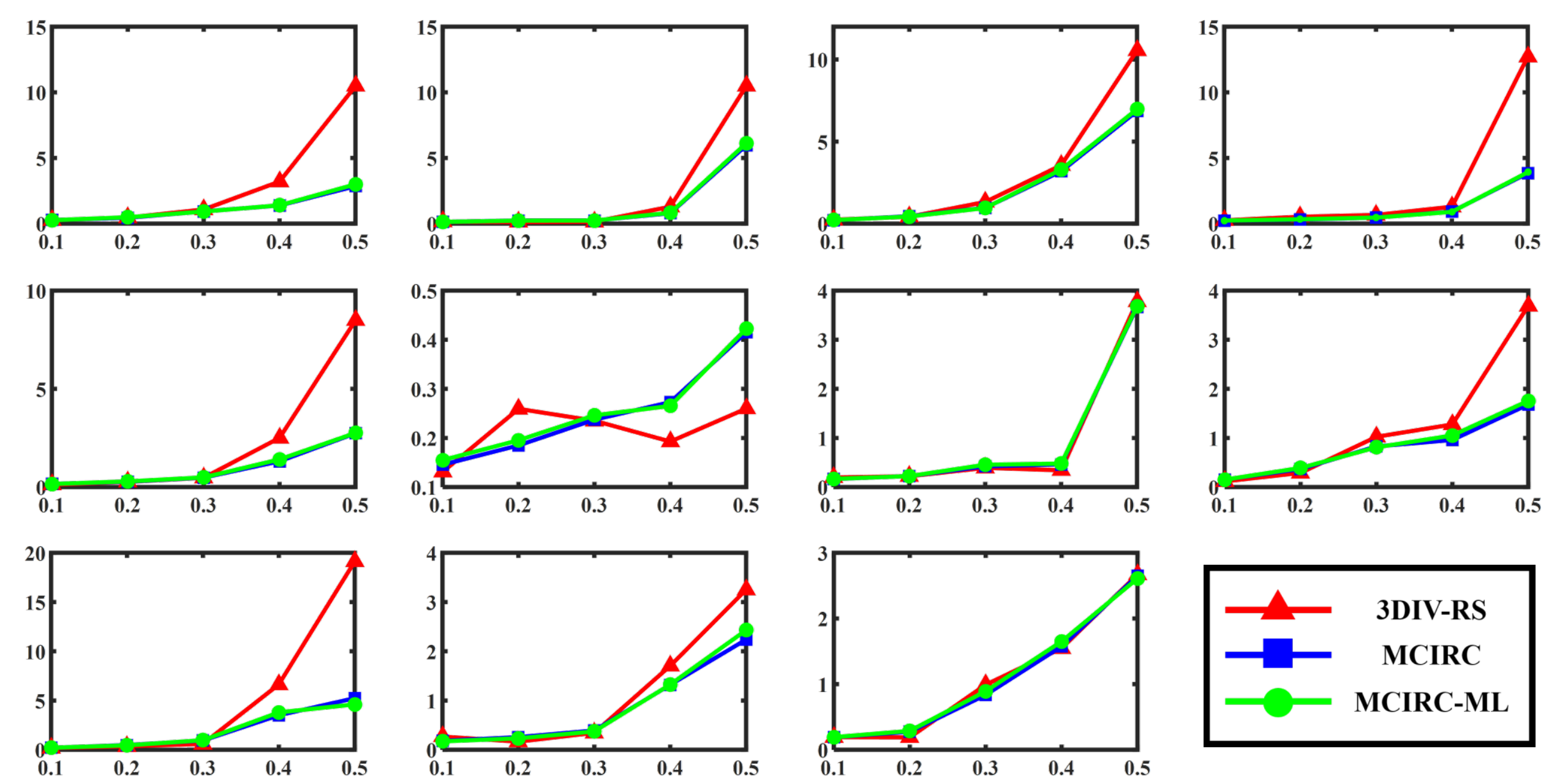}
\caption{Average runtime of 20 runs of 3DOF rotation search with respect to the missing points fraction for each model in the experiments with some data points deleted. The horizontal axis is the missing points fraction, and the vertical axis is the runtime (in seconds). From left top: Armadillo, Bunny, Camera, Chicken, Dragon, Hand, Happy Buddha, Rhino, T-rex, Asian dragon and Statuette. }
\end{figure}

\begin{figure}[tbp]
\centering
\includegraphics[width=3.6in]{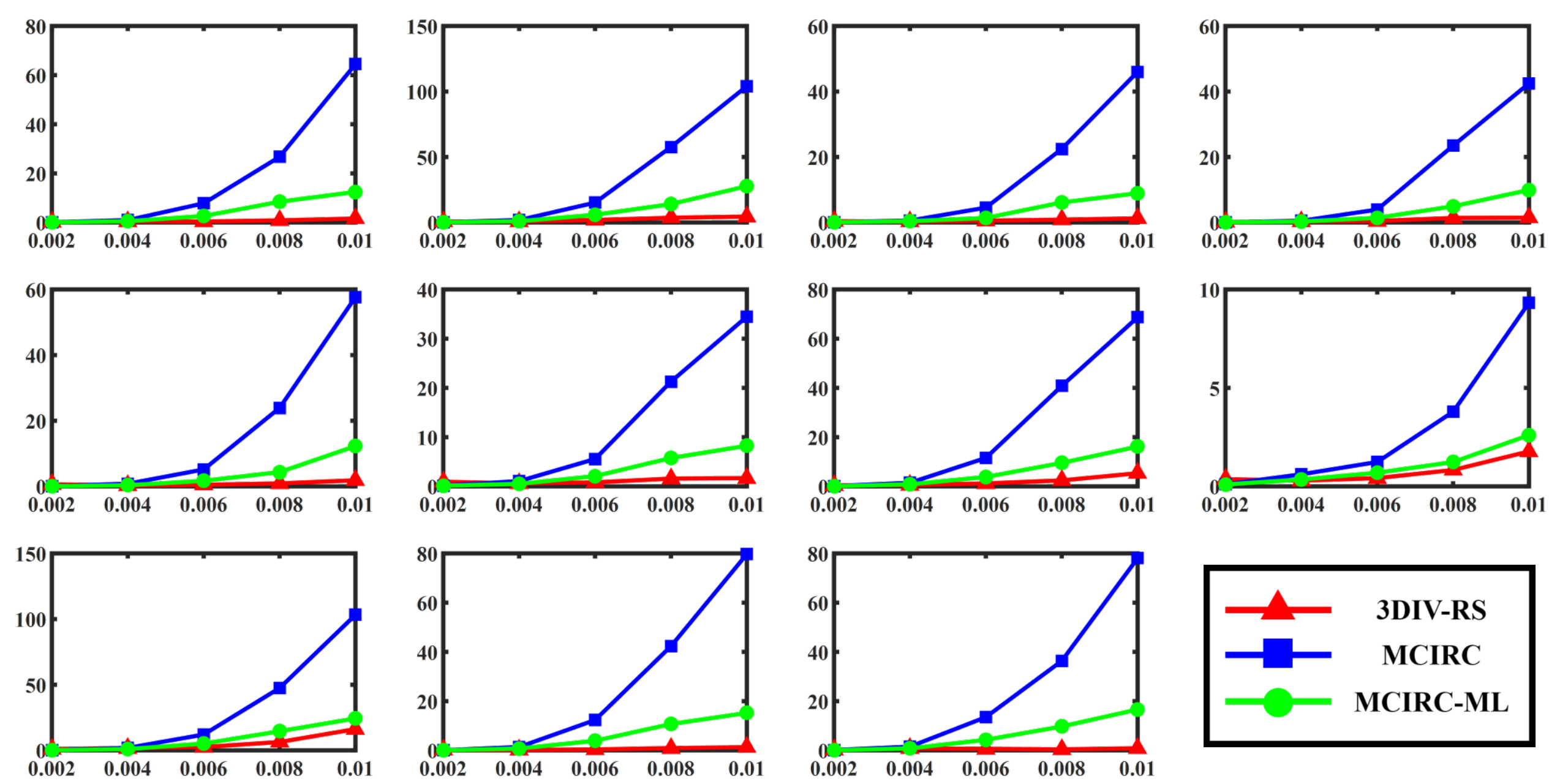}
\caption{Average runtime of 20 runs of 3DOF rotation search with respect to the noise level for each model in the experiments with perturbations of data points. The horizontal axis is the standard deviation of the Gaussian noise added to the points’ coordinates, and the vertical axis is the runtime (in seconds). From left top: Armadillo, Bunny, Camera, Chicken, Dragon, Hand, Happy Buddha, Rhino, T-rex, Asian dragon and Statuette. }
\end{figure}

In addition to the mean runtime, we also calculated the other three metrics in the 6-DOF experiments. The results are shown in Table 1, Table 2 and Table 3 for the three experimental situations, respectively. Note that APM (0.1) and Glob-GM-ML are not shown in the tables because they were unable to converge and terminate within the allotted 1,000 s in all the experiments. The proposed TIV-3DIV achieved successful registrations in all the experiments, while TIV-MCIRC, TIV-MCIRC-ML and Go-ICP experienced some failure cases, and APM (0.3) failed in all the experiments. 

\begin{table*}
\caption{The Results of The Experiments with Inserted Nondata Points}
\centering
\includegraphics[width=7in]{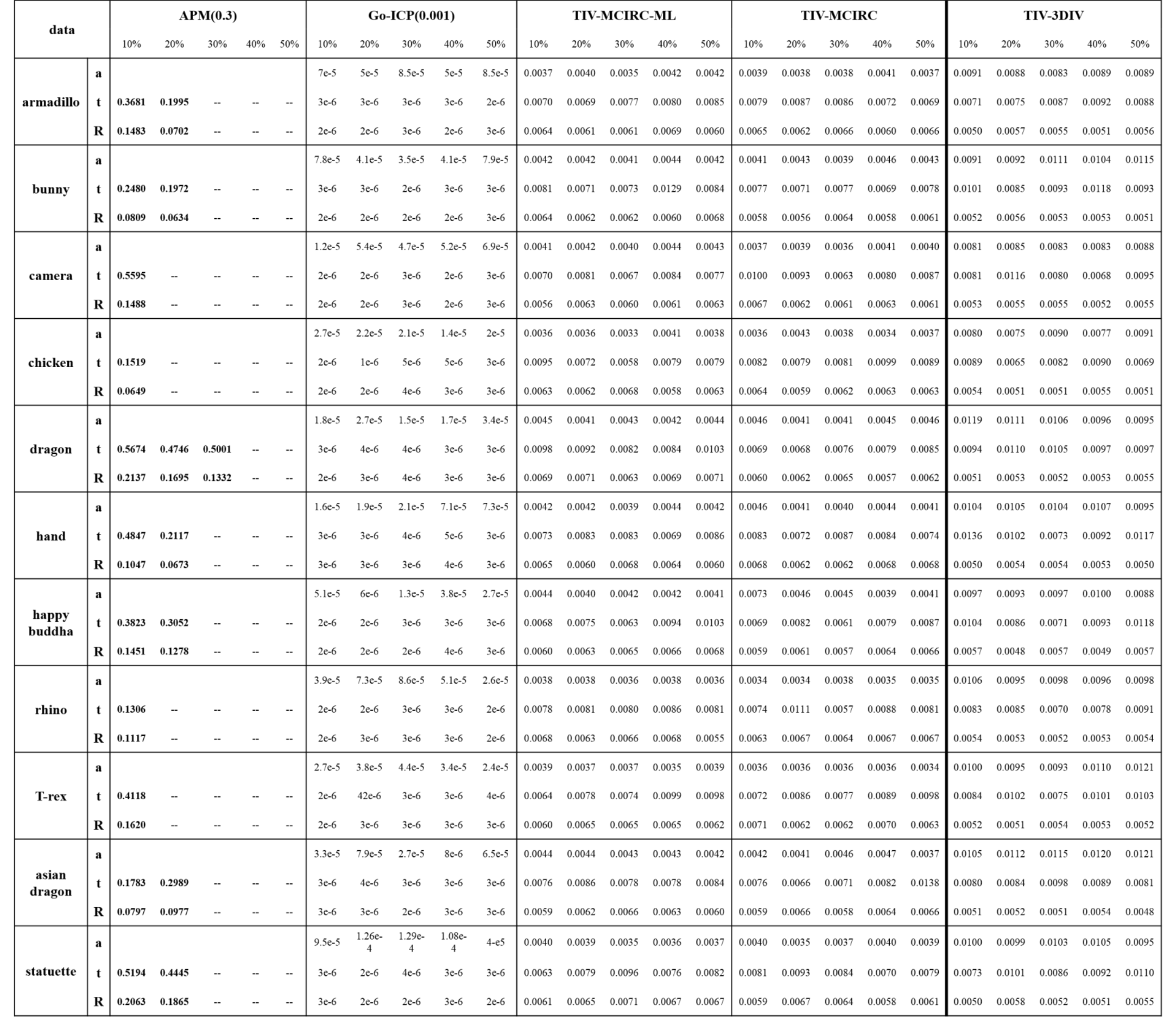}
\footnotemark{Here, a, t and R denote the angular error, translation error and RMS, respectively, and a "$--$" symbol means that the runtime exceeded the time limit (1,000 s). The bold numbers mean that among the 20 runs of random experiments, at least one failed to align the point clouds.}
\end{table*}

\begin{table*}
\caption{The Results of The Experiments with Some Data Points Deleted}
\centering
\includegraphics[width=7in]{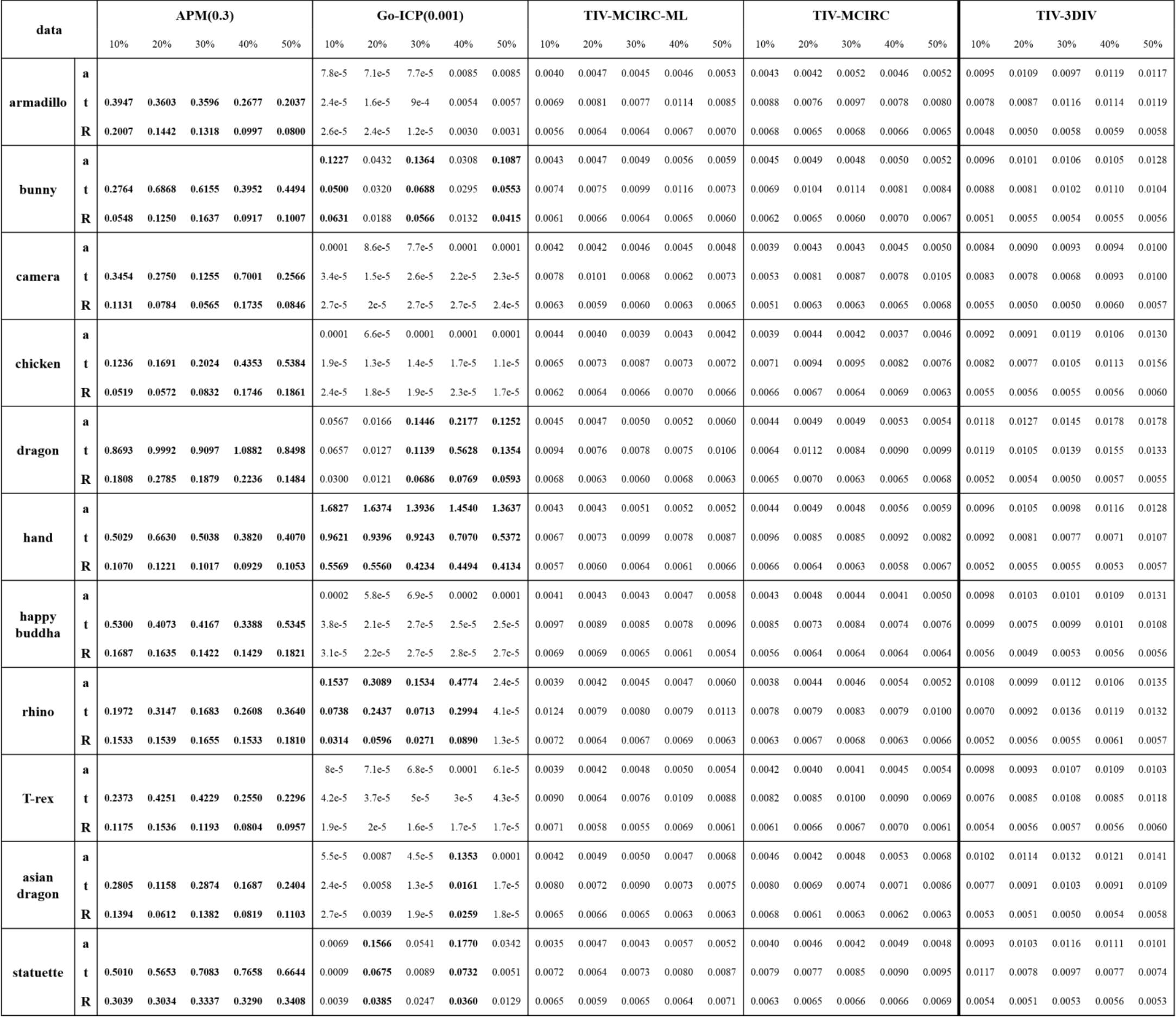}
\footnotemark{Here, a, t and R denote the angular error, translation error and RMS, respectively, and a "$--$" symbol means that the runtime exceeded the time limit (1,000 s). The bold numbers mean that among the 20 runs of random experiments, at least one failed to align the point clouds.}
\end{table*}

\begin{table*}
\caption{The Results of The Experiments with Perturbations of The Data Points}
\centering
\includegraphics[width=7in]{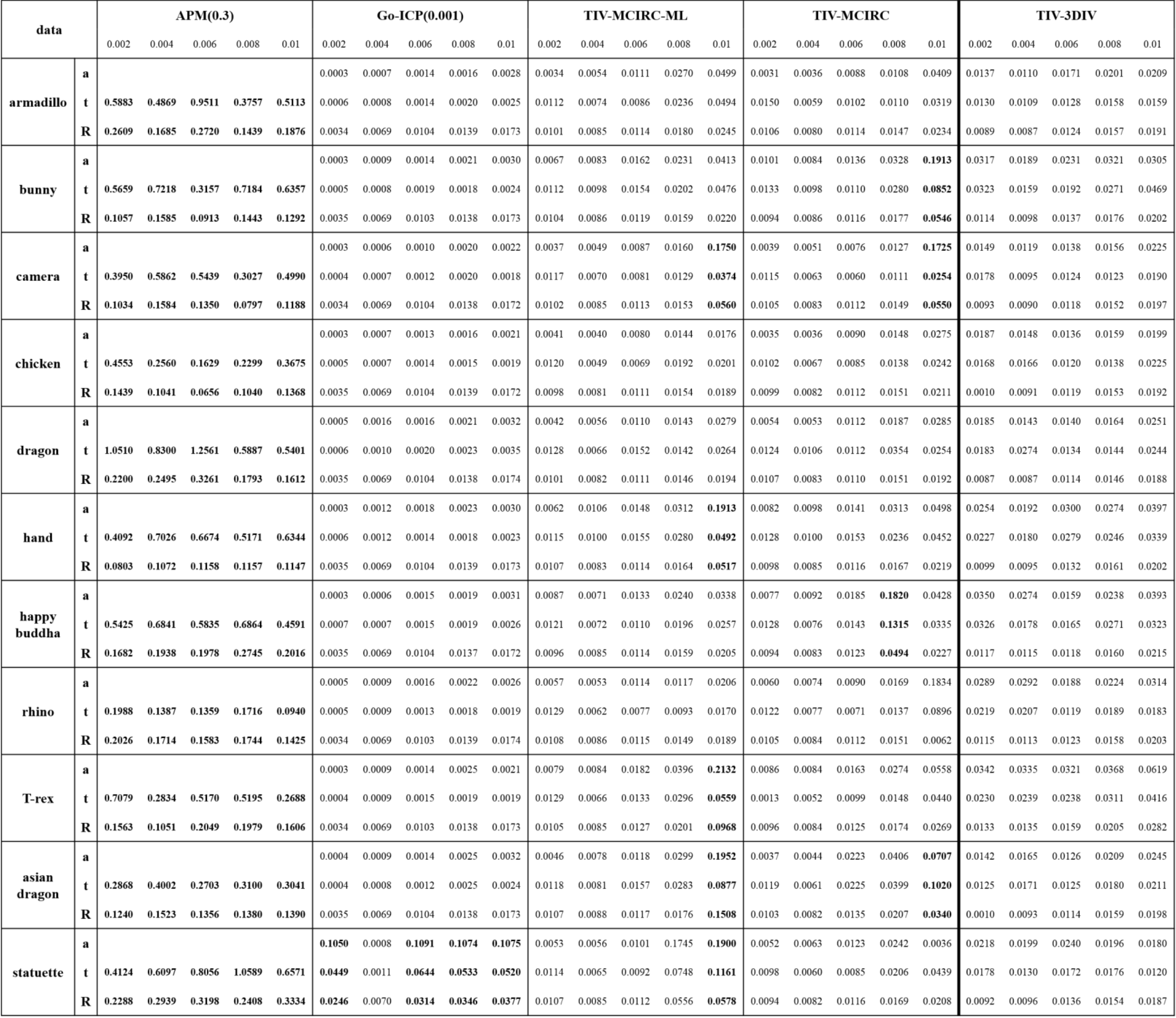}
\footnotemark{Here, a, t and R denote the angular error, translation error and RMS, respectively, and a "$--$" symbol means that the runtime exceeded the time limit (1,000 s). The bold numbers mean that among the 20 runs of random experiments, at least one failed to align the point clouds.}
\end{table*}

\subsection{The Scalability of TIV-3DIV}
In this experiment, we studied the changes in the TIV-3DIV runtime with respect to the number of points. We used the Hand model and down-sampled it to contain different numbers of points to generate the model and the scene set, using random relative rotation and translation. To study the scalability of the rotation kernel, we set the number of TIVs used for rotation search equal to 20$\%$ of the total number in the original point set rather than adopting the same TIV numbers used in the previous experiments. For each point number, 20 registrations were performed. The median times for each component of TIV-3DIV are illustrated in Fig. 13. From these results, we can see that the time required to construct a 3DIV is very small. The time required to construct and select TIVs is not negligible because it includes the time needed to sort the TIVs. The rotation and translation search times increase slowly as the number of points increases.

\begin{figure}[bp]
\centering
\includegraphics[width=3.5in]{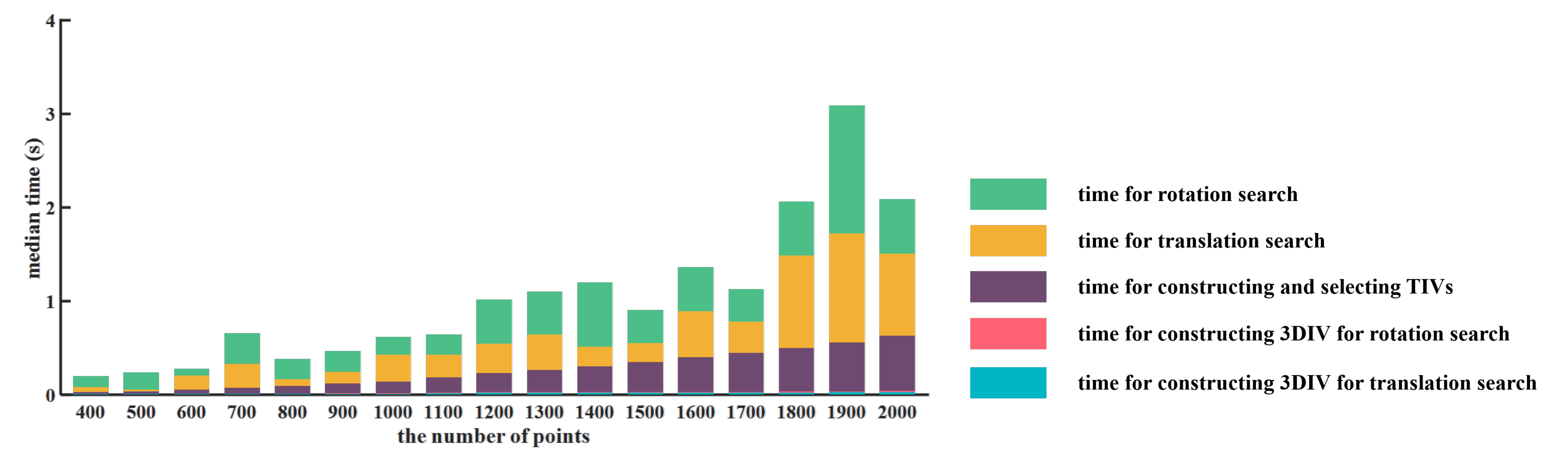}
\caption{Median run times of TIV-3DIV with respect to the number of points in the Hand model. For the point sets, which had different numbers of points, the number of selected TIVs was set to 20$\%$ of the number of points. }
\end{figure}

\subsection{Real Data}
In this section, we experimented on registrations with three sets of real data. Considering that APM optimizes the affine matrix instead of the rigid transformation and that the real data have no strict one-to-one correspondence, therefore, calculating RMS is invalid, APM was not included in the experiments in this part.

\subsubsection{Stanford Scanning Models}
First, we experimented on four Stanford scanning models [50]: Armadillo, Bunny, Dragon and Happy Buddha. Two partial scans obtained from different directions of each model were used in this experiment, and noise exists in the point coordinates. The gold standard values of relative rotation and translation between each pair of scans were provided in the dataset. The lack of a strict one-to-one correspondence and the partial overlap between scans of the same model lead to the presence of outliers. All the point sets were randomly down-sampled to 1,000 points, and uniform scaling was not employed here. For the three TIV-based methods, we deleted the 2,000 TIVs with the largest norms and selected the 500 TIVs with the largest norms among the remaining set. This selection strategy was used to make the selected TIVs robust to both outliers and noise. The 3DIV resolution remained the same as in Section 6.1, and the inlier thresholds were as follows: 0.004 for Armadillo, 0.0009 for Bunny, 0.0005 for Dragon and 0.004 for Happy Buddha. The implementation details of Go-ICP and Glob-GM-ML were the same as in Section 6.1. The results are presented in Table 4. Note that aligning the real data was more challenging; therefore, the time limit was set to 3,600 s for this experiment. Glob-GM-ML was still unable to complete the registration within the allotted time limit for all experiments, and TIV-MCIRC and TIV-MCIRC-ML were better in terms of both speed and accuracy. TIV-MCIRC-ML ran faster than TIV-MCIRC, which demonstrated that in difficult cases, the use of the matchlist technique is more significant. The Go-ICP (0.001) method failed to align Armadillo and Happy Buddha, and its angular errors were too large. TIV-3DIV achieved the fastest registration for all experiments, and its accuracy was among the best.
To show the optimality of TIV-3DIV, we present the evolutions of the upper and lower bounds of the rotation and translation searches in Fig. 14 for the experiments on Armadillo, Bunny, Dragon and Happy Buddha.

\begin{table*}
\caption{ The Results of Registration from Different Scans of Stanford Scanning Models}
\centering
\includegraphics[width=7in]{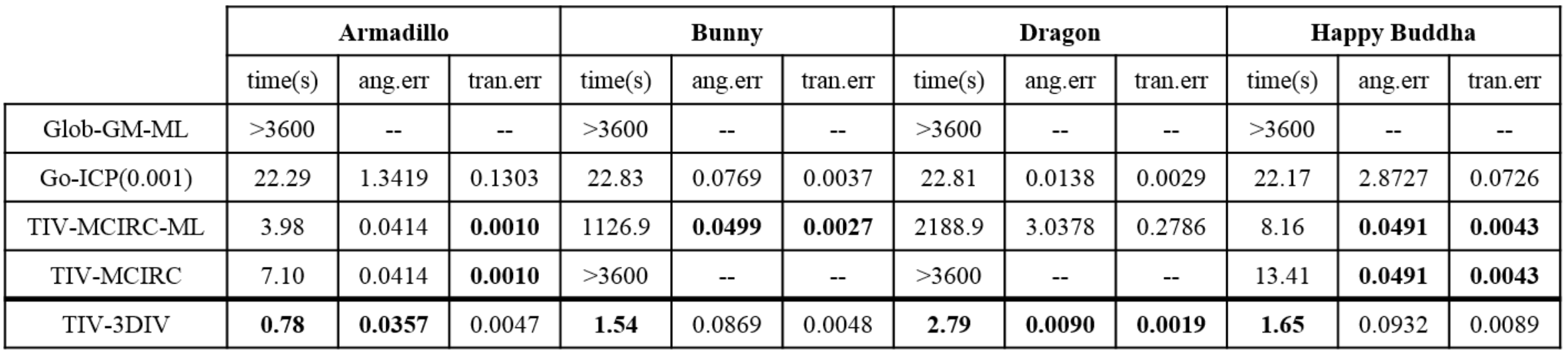}
\end{table*}

\begin{figure*}[htbp]
\centering
\subfloat[Armadillo]{\includegraphics[width=3.5in]{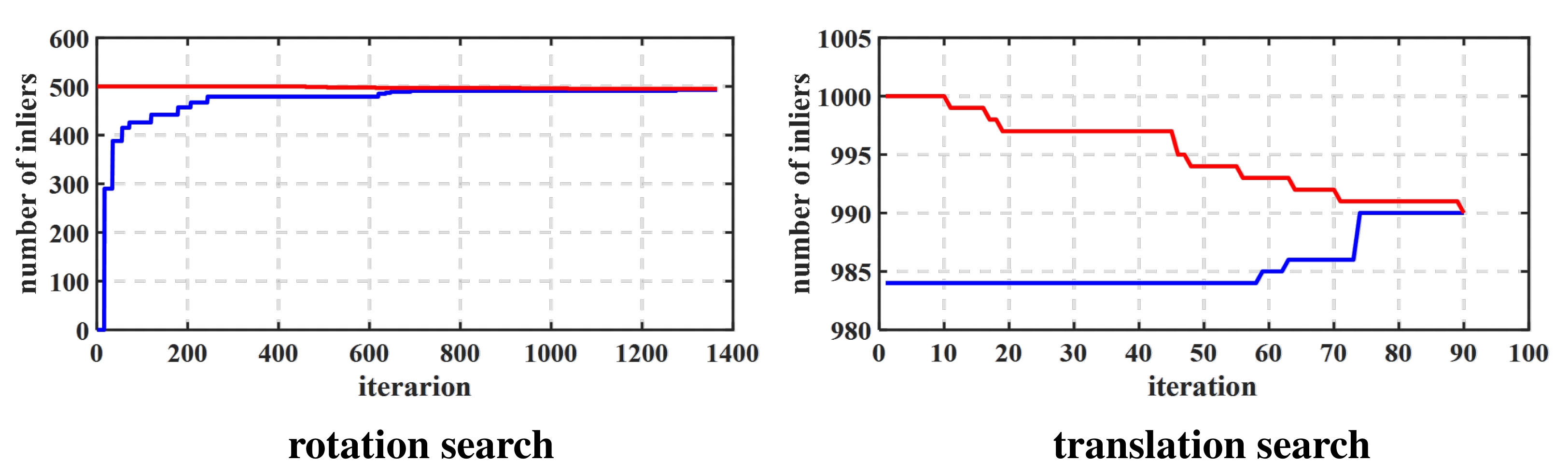}
\label{fig_first_case}}
\hfil
\subfloat[Bunny]{\includegraphics[width=3.5in]{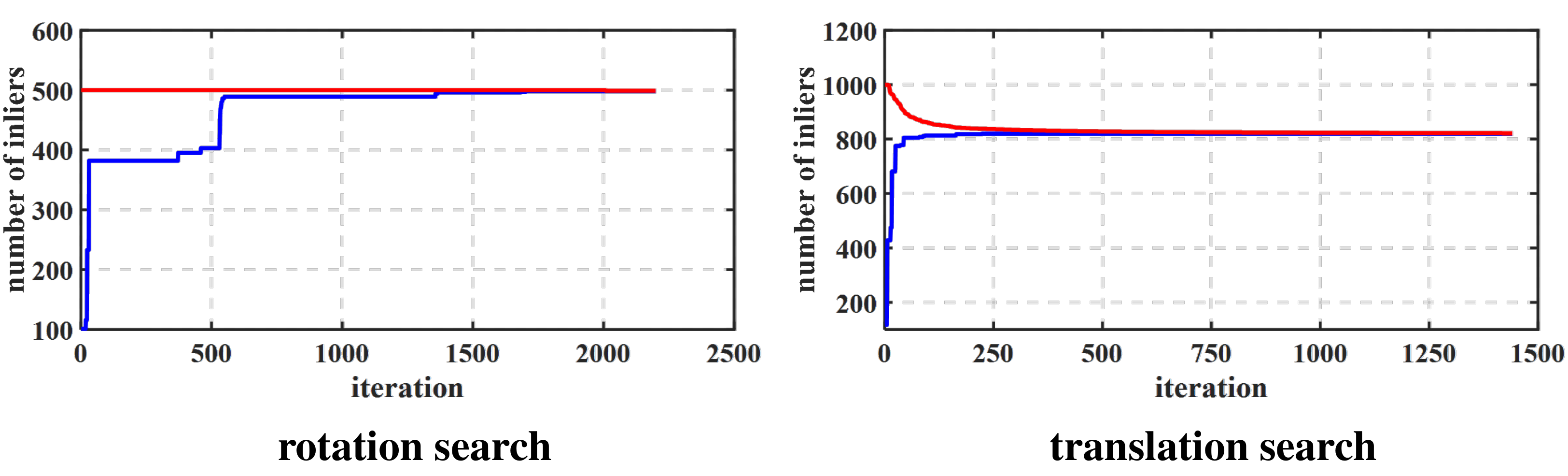}
\label{fig_second_case}}
\hfil
\subfloat[Dragon]{\includegraphics[width=3.5in]{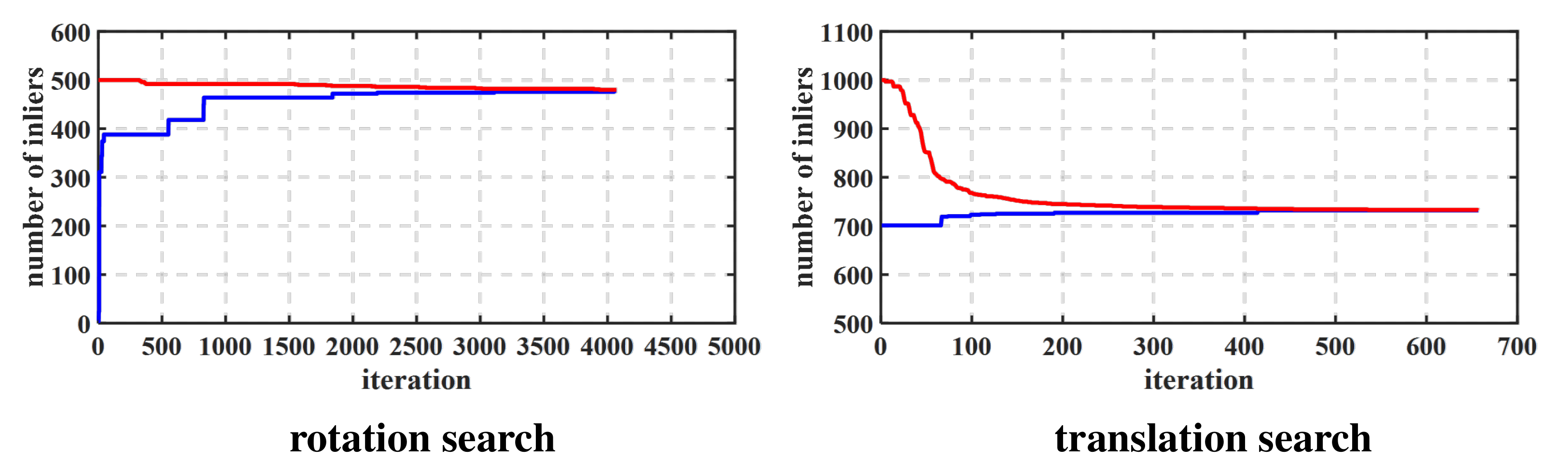}
\label{fig_third_case}}
\hfil
\subfloat[Happy Buddha]{\includegraphics[width=3.5in]{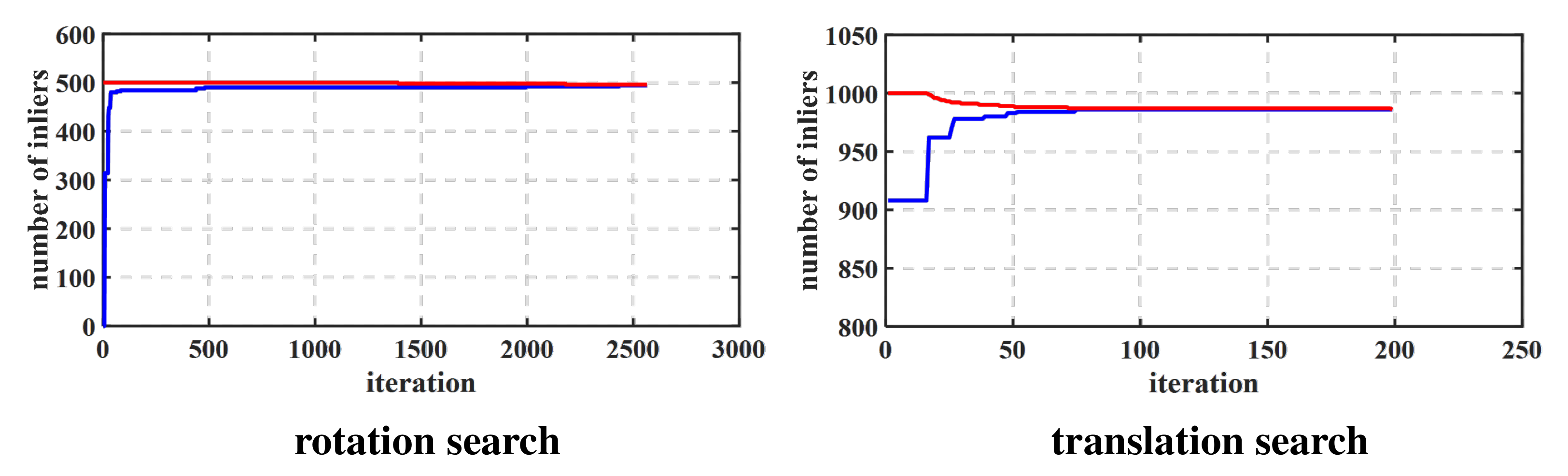}
\label{fig_fourth_case}}
\caption{The evolutions of the upper and the lower bounds of rotation search and translation search in the Armadillo, Bunny, Dragon and Happy Buddha experiments.}
\end{figure*}

\subsubsection{Indoor Scan Data}
Next, we tested these methods using a set of indoor scan data: the living room point clouds from MATLAB. These point clouds were captured using a Kinect from Microsoft. The two frames used in this experiment are shown in Fig. 15 (a). We randomly down-sampled the point clouds to 1,527 and 1,491 points. The 10,000 TIVs with the largest norms were deleted, and we selected 1,000 TIVs with the largest norms among the remaining set for the three TIV-based methods. The inlier threshold was 0.1 m, and the 3DIV resolution was  $51\times51\times51$. The implementation details of Go-ICP and Glob-GM-ML were the same as in Section 6.1 except that the convergence threshold of Go-ICP was 0.1 m. The results are listed in Table 5, and the point clouds before and after registration are shown in Fig. 15. The TIV-MCIRC and TIV-MCIRC-ML achieved the smallest translation error, while Go-ICP achieved the smallest angular error. TIV-3DIV was still the fastest method and its angular error and translation error are acceptable.

\subsubsection{Clinical Data}
Finally, we conducted a test on clinical data using spatial registration in image-guided neurosurgery [55]. Here, we registered a preoperative point cloud and an intraoperative point cloud to calculate a spatial transformation between the image space and the patient space. The preoperative point cloud was extracted from the 3D MRI volume data of the patient, while the intraoperative point cloud was obtained by scanning the patient’s head with a laser range scanner. The point clouds used in this experiment are shown in Fig. 16. Note that necessary preprocessing was conducted on the intraoperative point cloud (e.g., manual segmentation of the head area to avoid excessive searching caused by large outliers). The raw data preprocessing is both easy and common in practice; therefore, it does not compromise the results of our method. The processed data were randomly down-sampled to 1,000 points. The 2,000 TIVs with the largest norms were deleted, and we selected 500 TIVs with the largest norms among the remaining set for the three TIV-based methods. The inlier threshold was 5 mm, and the 3DIV resolution was  $51\times51\times51$. The implementation details of Go-ICP and Glob-GM-ML were the same as in Section 6.1 except that the convergence threshold of Go-ICP was 5 mm. We manually extracted the coordinates of four markers that which were adhered to the patient’s head before MRI scanning for point-based spatial registration from both point clouds; these formed the gold standard corresponding points for RMS calculation. The point clouds before and after registration are shown in Fig. 16, and the registration results are listed in Table 6. From Fig. 16(e), we can see that the original distance between the two point clouds is fairly extensive, and the relative rotation is also large. Obviously, this is a challenging registration. The corresponding points were extracted manually based on the marker points on the patients: this manual extraction may slightly influence the accuracy of the calculated RMS. Both Glob-GM-ML and Go-ICP failed to converge in 3,600 s; only the TIV-based methods were able to complete the registration within the time limit, but the errors for TIV-MCIRC and TIV-MCIRC-ML are too large. TIV-3DIV ranked first in terms of both runtime and accuracy on this challenging real-data task.

\begin{table}
\caption{The Registration Results of The Living Room Point Clouds}
\centering
\includegraphics[width=3.5in]{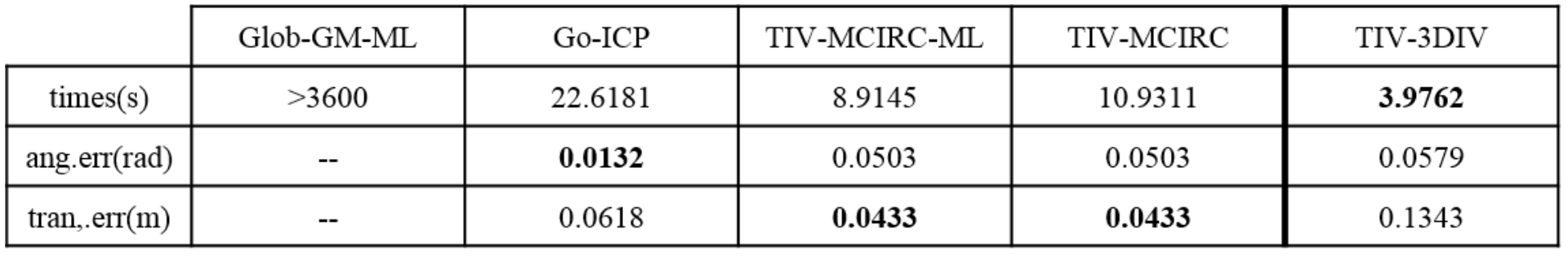}
\end{table}

\begin{figure}[tbp]
\centering
\includegraphics[width=3.5in]{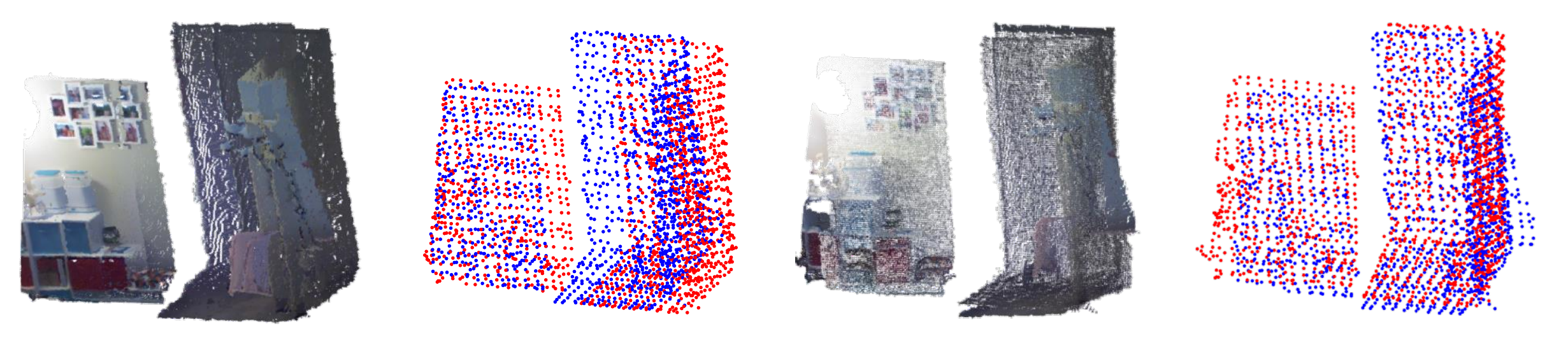}
\caption{ The living room point clouds before and after registration. From left to right: the initial scene point clouds, the initial down-sampled point clouds, the aligned and merged scene point clouds using TIV-3DIV, and the aligned down-sampled point clouds using TIV-3DIV.}
\end{figure}

\begin{figure}[tbp]
\centering
\subfloat[]{\includegraphics[width=0.7in]{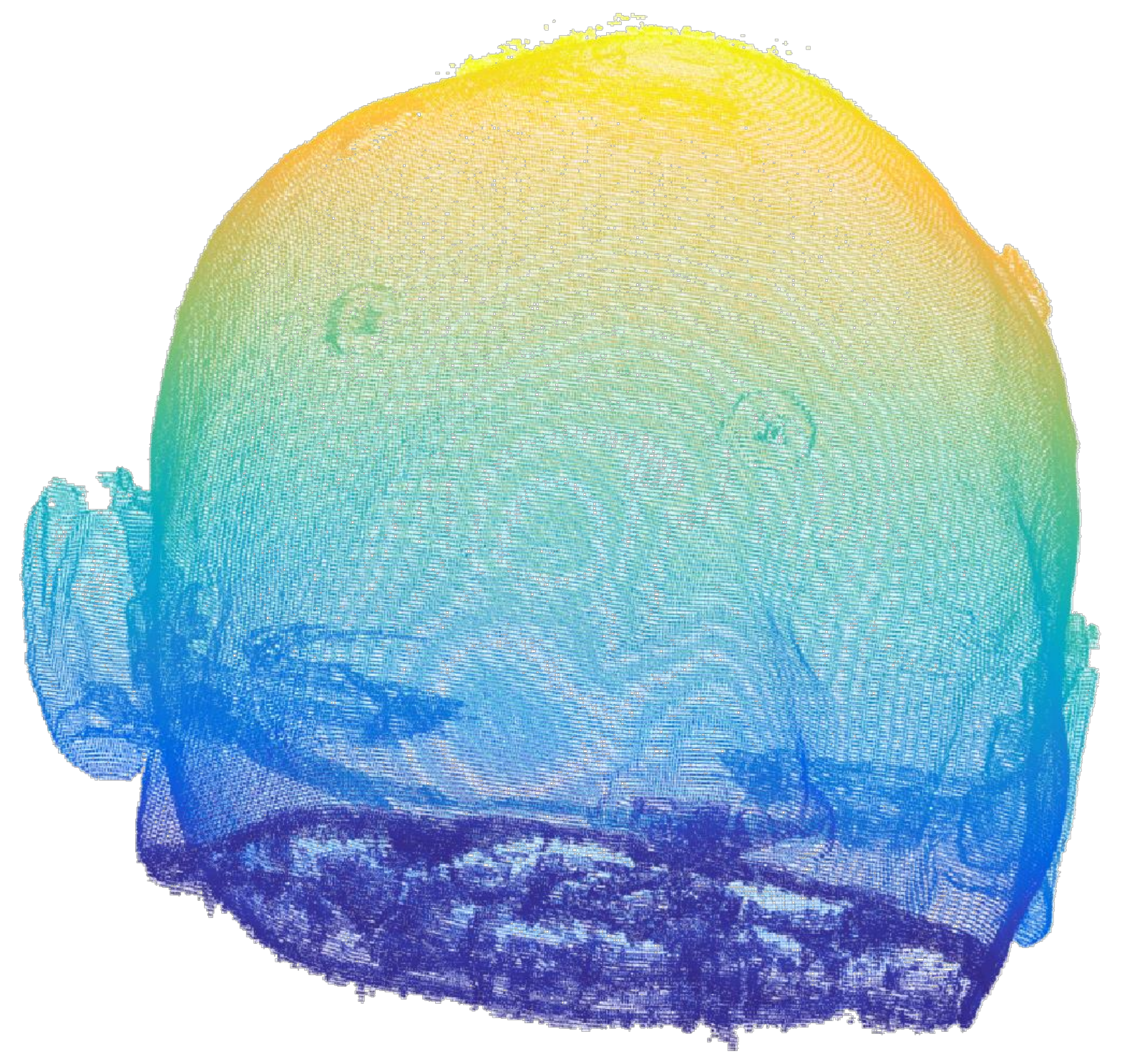}
\label{fig_first_case}}
\hfil
\subfloat[]{\includegraphics[width=0.8in]{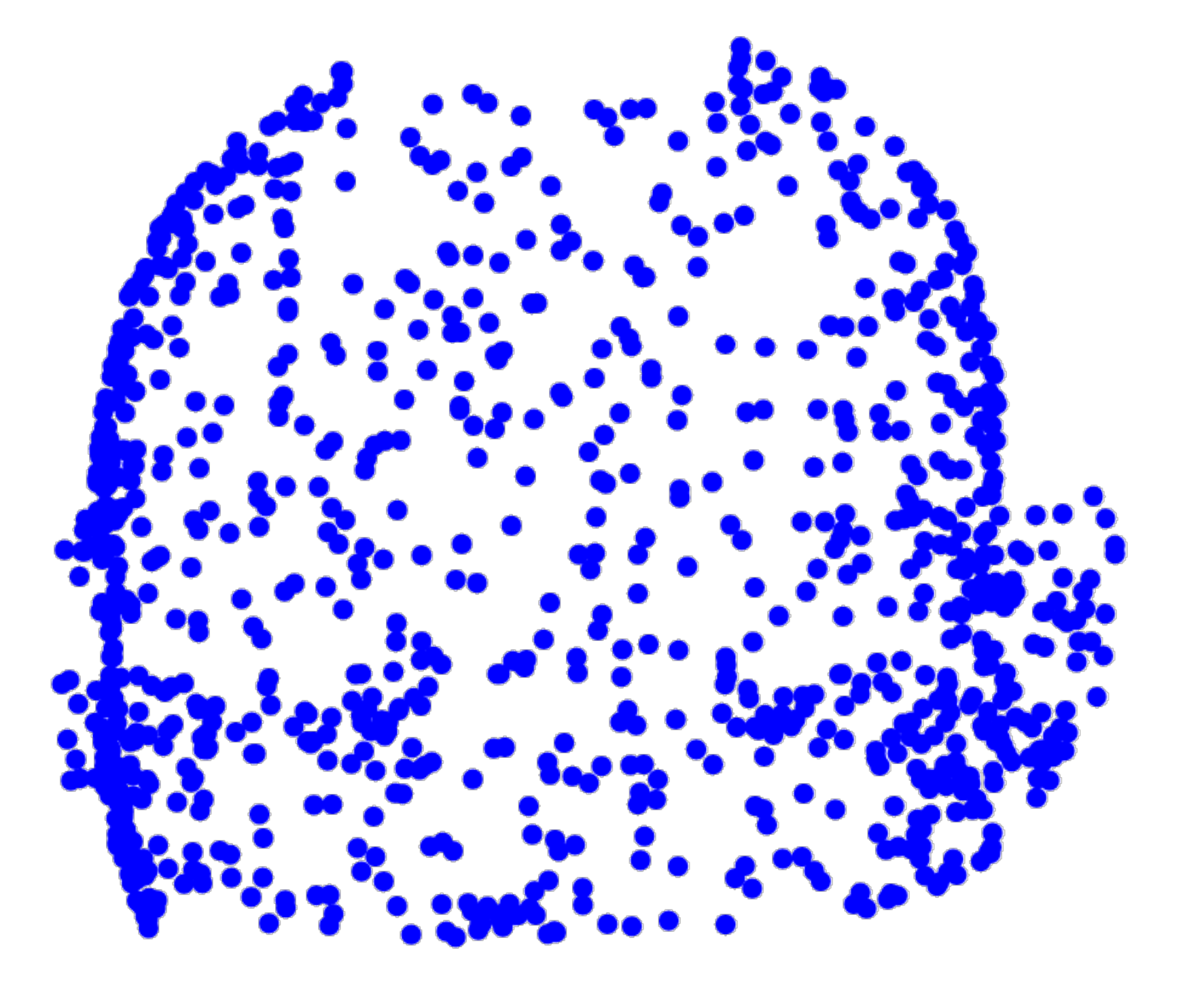}
\label{fig_second_case}}
\hfil
\subfloat[]{\includegraphics[width=0.6in]{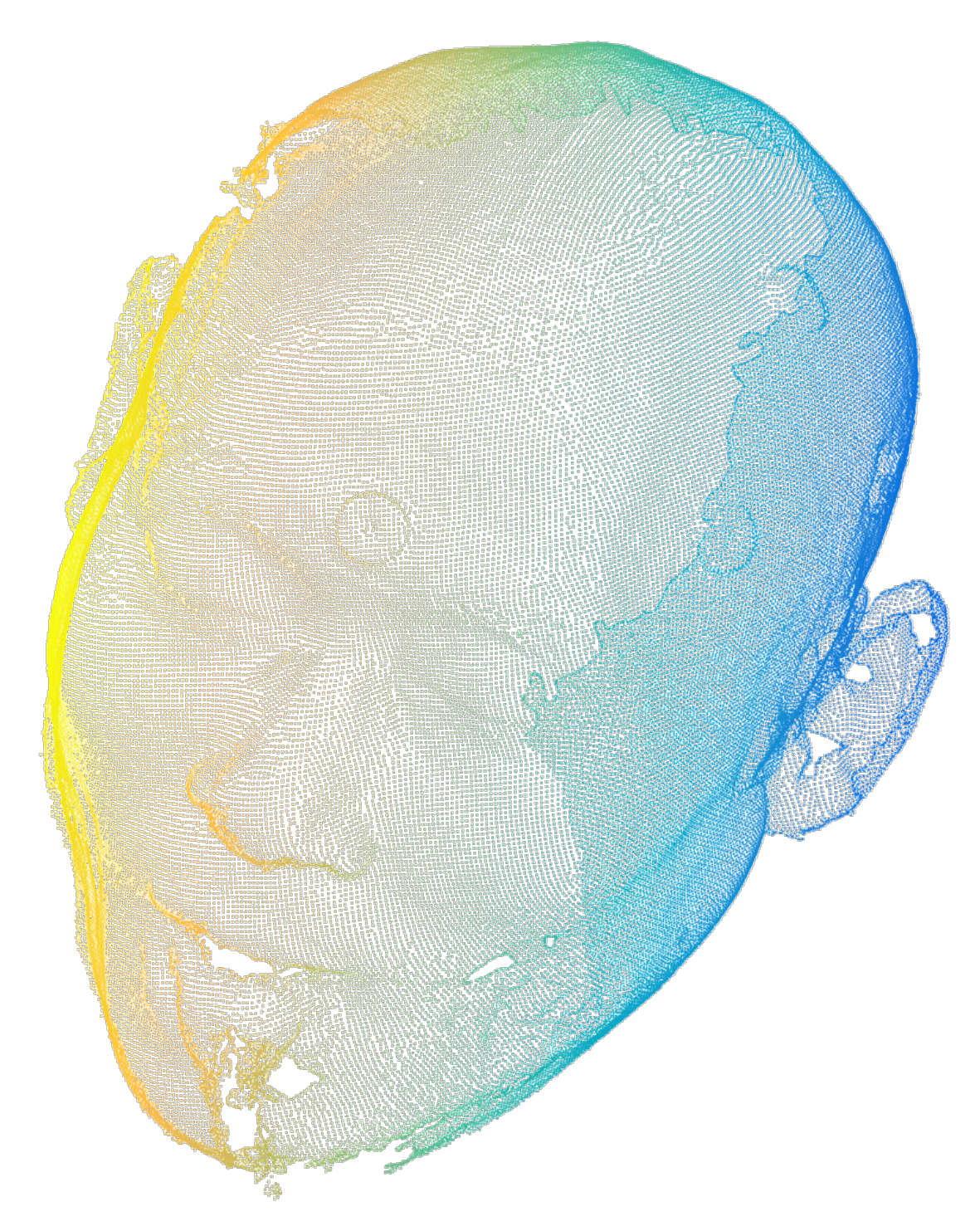}
\label{fig_third_case}}
\hfil
\subfloat[]{\includegraphics[width=0.8in]{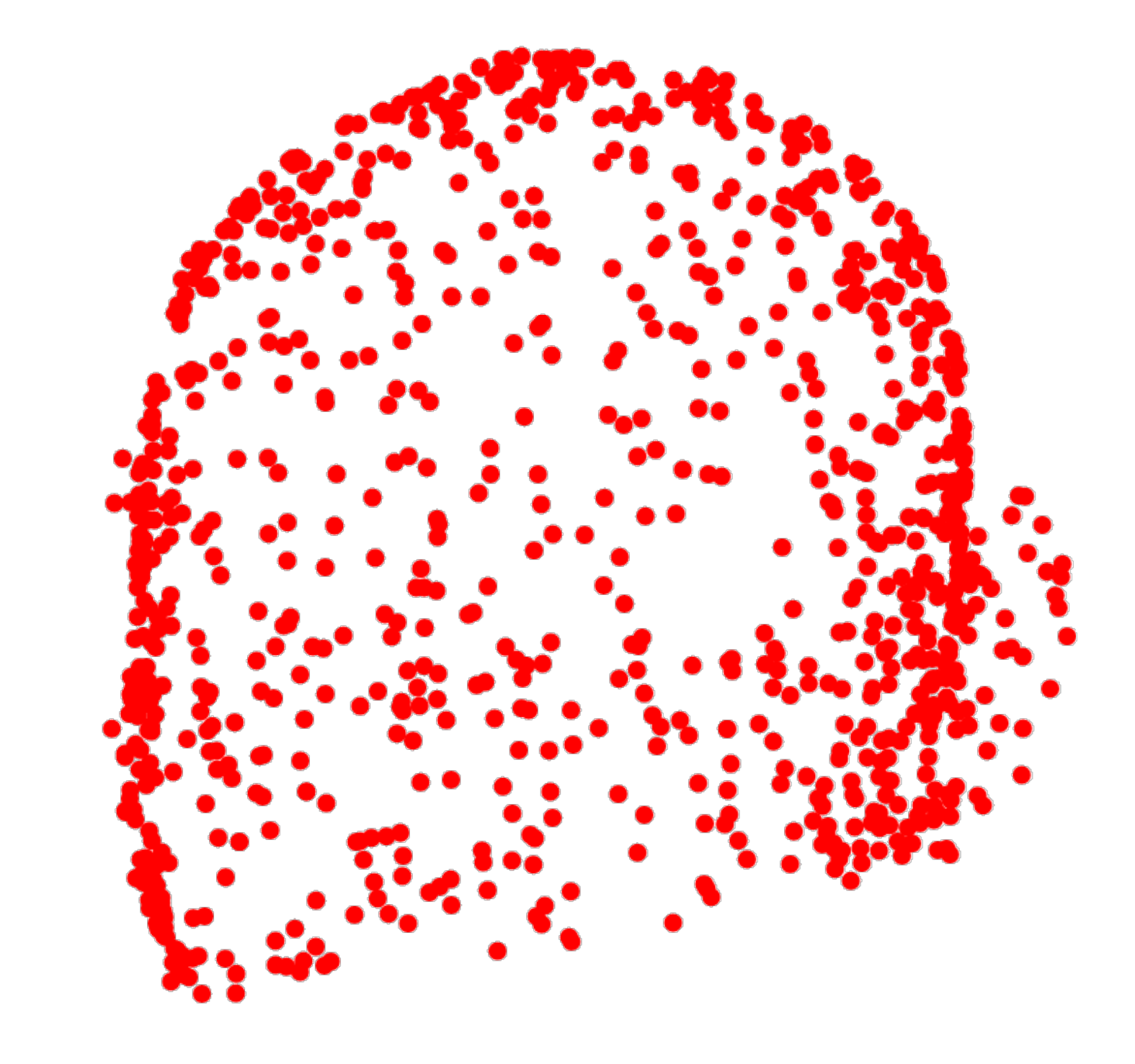}
\label{fig_fourth_case}}
\hfil
\subfloat[]{\includegraphics[width=2.5in]{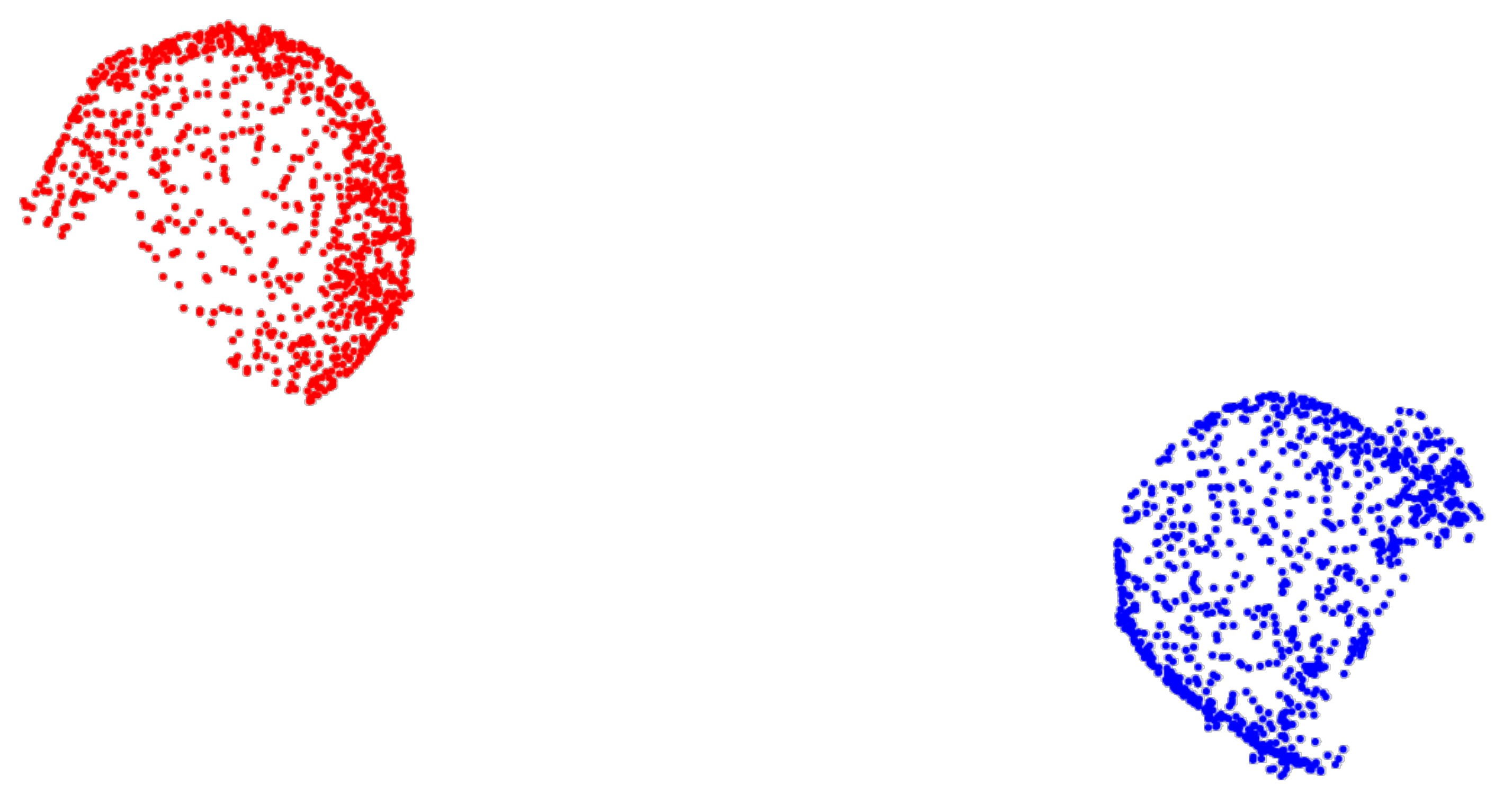}
\label{fig_fifth_case}}
\hfil
\subfloat[]{\includegraphics[width=0.8in]{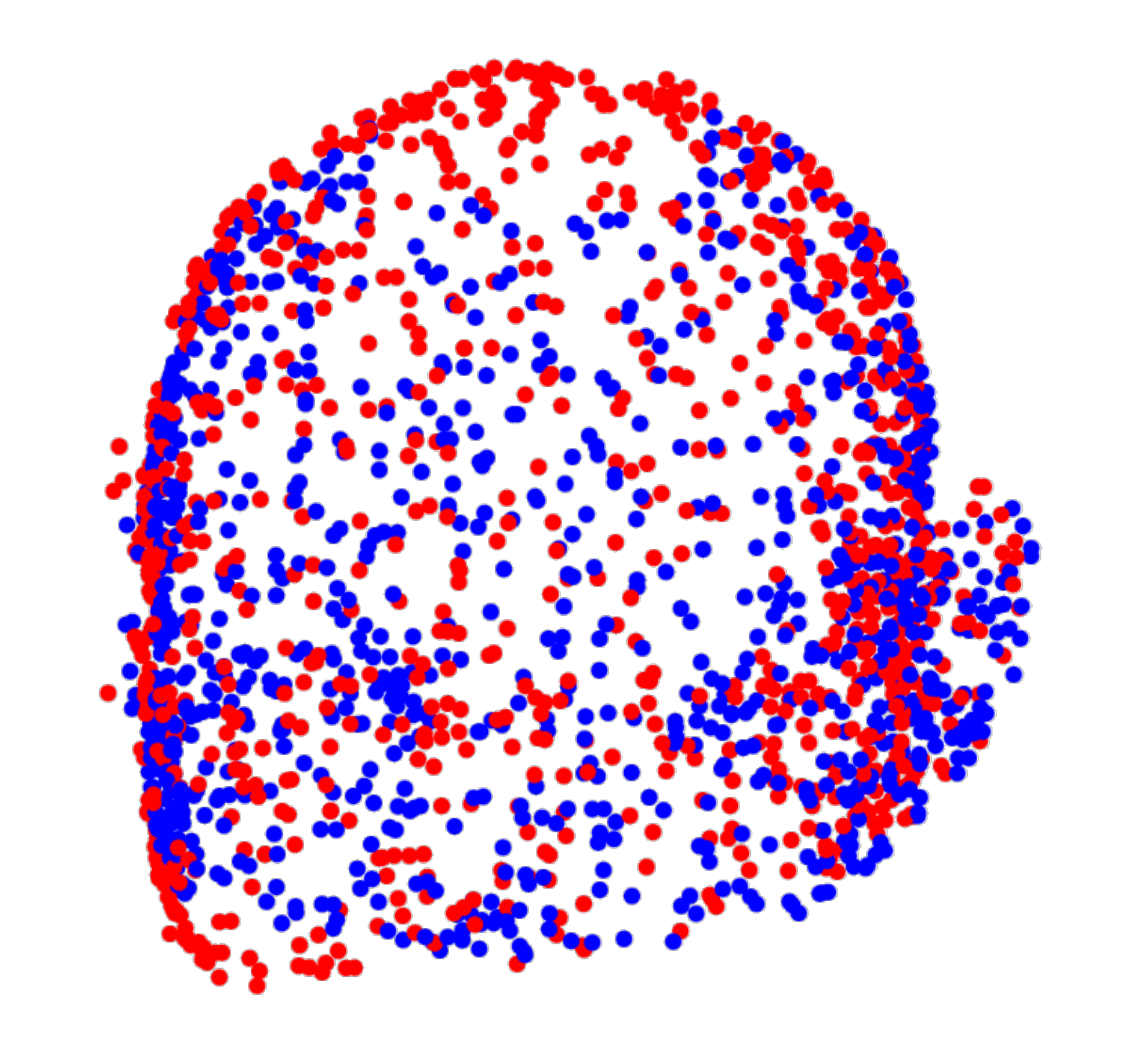}
\label{fig_sixth_case}}
\caption{An illustration of the registration of clinical data. First row: (a), (b) the original and preprocessed point sets extracted from 3D MRI volume data; (c), (d) the original and preprocessed point set scanned by a laser range scanner. The second row: (e) the initial relative positions of the two point sets to be registered; (f) the aligned point sets using TIV-3DIV.}
\end{figure}

\begin{table}
\caption{The Results of Registration on The Clinical Data}
\centering
\includegraphics[width=3.5in]{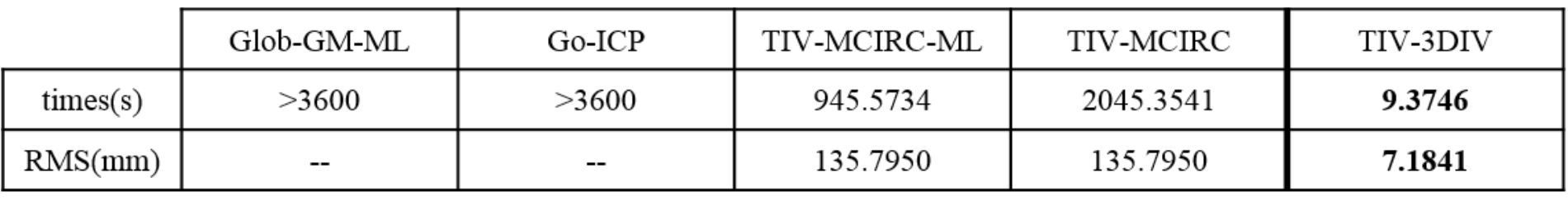}
\end{table}

\section{Conclusion}
In this paper, we propose a new efficient algorithm for global rigid registration of 3D point sets. The efficiency of this approach stems primarily from the proposed idea of transformation decomposition, in which we decompose the search of a 6D rigid transformation in the original problem into a 3D rotation search followed by 3D translation search. This reduction of the problem dimensionality greatly accelerates the BnB-based optimization algorithm. In addition, we propose using an  $L_\infty$-based consensus set as the objective function of registration and propose a new data structure, 3D Integral Volume, to speed up the evaluation of the bounds of the objective function in each branch of the parameter space. We conducted extensive experiments with both synthetic and real data to evaluate the performance of the proposed algorithm and compared it to existing state-of-the-art global methods. The results showed that the proposed method is three orders of magnitude faster than existing global methods and approximately ten times faster than Go-ICP, which is accelerated by a distance transform for fast bound evaluation.



\ifCLASSOPTIONcaptionsoff
  \newpage
\fi

\end{document}